\documentclass[runningheads]{llncs}
\usepackage{graphicx}
\usepackage{multirow}
\usepackage{amsfonts}
\usepackage{amsmath}
\usepackage{times}
\usepackage{graphicx}
\usepackage{blindtext}
\usepackage{amsmath}
\usepackage{amssymb}
\usepackage{csquotes}
\usepackage{comment}
\usepackage{hyperref}
\usepackage{textcomp}
\usepackage{multirow,fixltx2e}
\usepackage{etoolbox}
\patchcmd{\thebibliography}{\chapter*}{\section*}{}{}
\usepackage{array}
\usepackage[table,xcdraw]{xcolor}
\usepackage{multirow,fixltx2e}
\usepackage{microtype}
\usepackage{graphicx}
\usepackage{booktabs} 
\usepackage{algorithm}  
\usepackage{algpseudocode}
\usepackage{wrapfig,lipsum,booktabs}
\usepackage{graphicx}
\usepackage{caption}
\usepackage{subcaption,graphicx}
\usepackage[numbers]{natbib}
\usepackage{layouts}
\usepackage{subcaption}
\usepackage{subcaption}
\usepackage{algorithm}  
\usepackage{algpseudocode}
\usepackage{xcolor}
\usepackage{courier}  
\usepackage{array}
\newcolumntype{P}[1]{>{\centering\arraybackslash}p{#1}}
\newcolumntype{M}[1]{>{\centering\arraybackslash}m{#1}}

\usepackage{ragged2e}
\usepackage{csquotes}
\usepackage[T1]{fontenc}
\usepackage{graphicx}
\begin{document}
\title{Reward Delay Attacks on Deep Reinforcement Learning}
\titlerunning{Reward Delay Attacks on Deep Reinforcement Learning}
%
\author{Anindya Sarkar \and
Jiarui Feng \and
Yevgeniy Vorobeychik \and Christopher Gill \and Ning Zhang } 
\authorrunning{Sarkar et al.}
%
\institute{Washington University in St. Louis, MO 63130, USA \\
Department of Computer Science \& Engineering\\
\email{\{anindya,feng.jiarui,yvorobeychik,cdgill,zhang.ning\}@wustl.edu}}
\maketitle              
\begin{abstract}
Most reinforcement learning algorithms implicitly assume strong synchrony.
 We present novel attacks targeting Q-learning that exploit a vulnerability entailed by this assumption by delaying the reward signal for a limited time period.
 We consider two types of attack goals: \emph{targeted attacks}, which aim to cause a target policy to be learned, and \emph{untargeted attacks}, which simply aim to induce a policy with a low reward.
 We evaluate the efficacy of the proposed attacks through a series of experiments.
 Our first observation is that reward-delay attacks are extremely effective when the goal is simply to minimize reward. Indeed, we find that even naive baseline reward-delay attacks are also highly successful in minimizing the reward.  
 Targeted attacks, on the other hand, are more challenging, although we nevertheless demonstrate that the proposed approaches remain highly effective at achieving the attacker's targets.
 In addition, we introduce a second threat model that captures a minimal mitigation that ensures that rewards cannot be used out of sequence.
 We find that this mitigation remains insufficient to ensure robustness to attacks that delay, but preserve the order, of rewards.

\end{abstract}
\section{Introduction}
In recent years, deep reinforcement learning (DRL) has achieved super-human level performance in a number of applications including game playing~\citep{silver2018general}, clinical decision support~\citep{liu2020reinforcement}, and autonomous driving~\citep{kiran2021deep}.
However, as we aspire to bring DRL to safety-critical settings, such as autonomous driving, it is important to ensure that we can reliably train policies in realistic scenarios, such as on autonomous vehicle testing tracks~\citep{ACM22,dong2019mcity,STII22}.
In such settings, reward signals are often not given exogenously, but derived from sensory information.
For example, in lane following, the reward may be a function of vehicle orientation and position relative to the center of the lane, and these features are obtained from perception~\cite{garnett20193d,wang2018lanenet}.
Since many such settings are also safety-critical, any adversarial tampering with the training process---particularly, with the integrity of the reward stream derived from perceptual information---can have disastrous consequences.

A number of recent efforts demonstrated vulnerability of deep reinforcement learning algorithms to adversarial manipulation of the reward stream~\citep{behzadan2017vulnerability,huang2017adversarial,kos2017delving,lin2017tactics}.
We consider an orthogonal attack vector which presumes that the adversary has compromised the \emph{scheduler} and is thereby able to manipulate reward timing, but cannot modify rewards directly.
For example, ROS 2.0 features modular design with few security checks and the ability to substitute different \emph{executors}~\citep{Casini2019RTAROS2,Tang2020RTAROS2,Blass2021ALMROS2,Choi2021PiCASROS2,Blass2021RTAROS2}.
This means that once adversaries gain access to the ROS software stack, they can replace its scheduling policy readily, and as long as the executor behavior is not overtly malicious it can be a long time before the compromise has been discovered.
Additionally, we assume that the adversary  can infer (but not modify) memory contents using side channel attacks. This is a realistic assumption, since it has been demonstrated that it is feasible to leverage different types of system or architectural side channels, such as cache-based or proc-fs based side channels to infer secret information in other ROS modules~\citep{luo2020stealthy,ravensec_raid2019,kocher2019spectre}. However, write access to memory is often a lot more difficult to obtain due to existing process isolation~\cite{dieber2016application,demarinis2019scanning}.    

Our attack exploits a common assumption of \emph{synchrony} in reinforcement learning algorithms.
Specifically, we assume that the adversary
can delay rewards a bounded number of time steps (for example, by scheduling tasks computing a reward at time $t$ after the task computing a reward at time $t+k$ for some integer $k\ge 0$).
We consider two variations of such \emph{reward delay} attacks.
In the first, we allow the adversary to arbitrarily shuffle or drop rewards, assuming effectively that no security mechanisms are in place at all.
Our second model evaluates the efficacy of the most basic security mechanism in which we can detect any rewards computed out of their arrival sequence, for example, through secure time stamping.
Consequently, we propose the \emph{reward shifting} attacks, where in order to remain undetected, the adversary can only drop rewards, or \emph{shift} these a bounded number of steps into the future. 
Efficacy comparison between these two threat models will then exhibit the extent to which this simple security solution reduces vulnerability to reward delay attacks.
In both attack variants, we consider two adversarial goals: untargeted attacks, which aim to minimize total reward accumulated at prediction time (essentially, eroding the efficacy of training), and targeted attacks, the goal of which is to cause the RL algorithm to learn a policy that takes target actions in specific target states.



We specifically study attacks on deep Q-learning algorithms.
The adversarial model we introduce is a complex discrete dynamic optimization problem, even in this more narrow class of DRL algorithms.
We propose an algorithmic framework for attacks that is itself based on deep Q learning, leveraging the fact that the current Q function, along with the recent sequence of states, actions, and rewards observed at training, provide sufficient information about system state from the attacker's perspective.
The key practical challenge is how to design an appropriate reward function, given that the ``true'' reward is a property of the final policy resulting from long training, and only truly possible to evaluate at test time.
We address this problem by designing proxy reward functions for both untargeted and targeted attacks that make use of only the information immediately available at training time.

We evaluate the proposed approach experimentally using two Atari games in OpenAI gym: Pong and Breakout.
Our experiments demonstrate that the proposed attacks are highly effective, and remain nearly as effective even with the simple mitigation that ensures that rewards are not encountered out of order.
Altogether, our results demonstrate the importance of implementing sound security practices, such as hardware and software-level synchrony assurance~\citep{li2021chronos,mahfouzi2019butterfly}, in safety-critical applications of reinforcement learning.


\section{Related Work}
There are two closely related literature strands in attacks on reinforcement learning and multiarmed bandits: attacks that take place at decision time, and poisoning attacks.

\paragraph{Decision-Time attacks on Reinforcement Learning}
Prior literature on adversarial attacks against RL has focused mainly on inference time attacks  \citep{behzadan2017vulnerability,huang2017adversarial,kos2017delving,lin2017tactics}, where the RL policy $\pi$ is pre-trained and fixed, and the attacker manipulates the perceived state $s_t $ of the learner to $s^{\prime}_{t}$ in order to induce undesired actions, while restricting $s^{\prime}_{t}$ and $s_t$ to be very similar to (or human-indistinguishable from) each other. For example, in video games the attacker can make small pixel perturbations to a frame to induce an action $\pi(s^{\prime}_{t}) \neq \pi(s_t)$. \cite{huang2017adversarial} developed the uniform attack mechanism, which generates adversarial examples by perturbing each image the agent observes to attack a deep RL agent at every time step in an episode in order to reduce the agent's reward. \cite{huang2017adversarial} also introduced a decision time targeted attack strategy, i.e. enchanting attack tactic, which is a planning-based adversarial attack to mislead the agent towards a target state. \cite{lin2017tactics} proposed a strategically timed attack, which can reach the same effect of the uniform attack by attacking the agent four times less often on average. \cite{kos2017delving} leverages the policy’s value function as a guide for when to inject adversarial perturbations and shows that with guided injection, the attacker can inject perturbations in a fraction of the frames, and this is more successful than injecting perturbations with the same frequency but no guidance. \cite{behzadan2017vulnerability} proposes an attack mechanism that exploits the transferability of adversarial examples to implement policy induction attacks on deep Q networks. Although test-time attacks can severely affect the performance of a fixed policy $\pi$ during deployment, they do not modify the policy $\pi$ itself.

\paragraph{Reward Poisoning Attacks on Reinforcement Learning}
Reward poisoning has been studied in bandits \cite{altschuler2019best,jun2018adversarial,ma2018data,liu2019data}, where the authors show that an adversarially perturbed reward can mislead standard bandit algorithms to suffer large regret. Reward poisoning has also been studied in batch RL where rewards are stored in a pre-collected batch data set by some behavior policy, and the attacker modifies the batch data \cite{huang2019deceptive,zhang2009policy,zhang2008value,zhang2020adaptive}. \cite{huang2019deceptive} provides a set of threat models for RL and establishes a framework for studying strategic manipulation of cost signals in RL. \cite{huang2019deceptive} also provides results to understand how manipulations of cost signals can affect Q-factors and hence the policies learned by RL agents. \cite{zhang2020adaptive} proposed an adaptive reward poisoning attack against RL, where the perturbation at time t not only depends on $(s_t,a_t,s_{t+1})$ but also relies on the RL agent's Q-value at time t. \cite{zhang2009policy} presents a solution to the problem of finding limited incentives to induce a particular target policy, and provides tractable methods to elicit the desired policy after a few interactions. Note that all these previous works directly modify the value of the reward signal itself, by adding a quantity $\delta_t$ to the true reward $r(t)$, i.e.  $r^{\prime}(t) = r_t + \delta_{t}$. In contrast, we focus on delaying the reward signal with the aim to mislead a learner RL agent, but cannot directly modify the rewards. 
As such, our key contribution is the novel threat model that effectively exploits the common synchrony assumption in reinforcement learning.

\section{Model}

Consider a discounted Markov Decision Process (MDP) with a set of states $S$, set of actions $A$, expected reward function $r(s,a)$, transition function $P_{ss'}^\alpha = \Pr\{s_{t+1} = s'|s_t = s,a_t = a\}$, discount factor $\gamma \in [0,1)$, and initial state distribution $D(s) = \Pr\{s_0 = s\}$.
Suppose that we only know $S$ and $A$, but must learn an optimal policy $\pi(s)$ from experience using reinforcement learning.
To this end, we consider a Deep $Q$-Network (DQN) reinforcement learning framework; it is straightforward to extend our approach to other variants of deep $Q$ learning.
Let $Q(s,a;\theta)$ be a neural network with parameters $\theta$ representing the $Q$-function, and let $Q_t(s,a;\theta_t)$ denote an approximation of the $Q$-function at iteration $t$ of RL (which we also denote by $Q_t$ when the input is clear, and $Q_t(s,a;\theta)$ when we treat $\theta$ as  a variable). In the DQN, 
parameters $\theta$ are updated after each iteration using the loss function
\(
L(\theta) = (r_t + \gamma \max_{a'} Q_t(s_{t+1},a';\theta) - Q_t(s_t,a_t;\theta))^2.
\)
Below, we omit the explicit dependence on $\theta$.
As this update rule makes evident, without experience replay the DQN training process itself is a Markov decision process (MDP) in which $\sigma_t = (Q_t,(s_t,a_t,r_t,s_{t+1}))$ constitutes state.
Let $T$ be the total number of DQN update iterations.
This observation will be useful below.
Given a Q function obtained at the end of $T$ learning iterations, $Q_T(s,a)$, we assume that the learner will follow a deterministic policy that is optimal with respect to this function, i.e., $\pi(s) = \arg\max_a Q_T(s,a)$.
For convenience, we abuse this notation slightly, using $\pi(s,a)$ as an indicator which is 1 if action $a$ is played in state $s$, and 0 otherwise.

Suppose that the attacker has compromised the scheduler, which can delay a reward computed at any time step by a bounded number of time steps $\delta$ (for example, to prevent attacks from appearing too obvious).  
Attacks of this kind take advantage of the settings in which reward needs to be computed based on perceptual information.
For example, the goal may be to learn a lane-following policy, with rewards computed based on vehicle location relative to lane markers inferred from camera and GPS/IMU data.
A compromised scheduler can delay 
the computation associated with a reward, but cannot directly modify rewards (contrasting our attacks from prior research on reward poisoning \cite{huang2019deceptive,zhang2009policy,zhang2008value,zhang2020adaptive}).
As a useful construct, we endow the attacker with a $\delta$-sized disk $\mathcal{D}$ in which the past rewards are stored, and the attacker can utilize $\mathcal{D}$ to replace the original reward used to update the $Q$-function parameters at time $t$.

We consider two common attack goals: 1) \emph{untargeted attacks} which simply aim to minimize the reward obtained by the learned policy, and 2) \emph{targeted attacks}, which attempt to cause the learner to learn a policy that takes particular target actions.
To formalize these attack goals, let $\mathbf{r}^\alpha = \{r_0^\alpha,\ldots,r_{T-1}^\alpha\}$ be an adversarial reward stream induced by our attack, resulting in the learned Q function $\tilde{Q}_T(s,a;\mathbf{r}^\alpha)$.
On the other hand, let $Q_T(s,a)$ be the Q function learned without adversarial reward perturbations.
Let $\tilde{\pi}(s;\mathbf{r}^\alpha)$ be the policy induced by $\tilde{Q}_T$, with $\tilde{\pi}(s,a;\mathbf{r}^\alpha)$ a binary indicator of which action is taken in which state.
The goal of an untargeted attack is 
\begin{equation}
\label{E:untargeted}
\min_{\mathbf{r}^\alpha} \sum_{a} \tilde{\pi}(s,a;\mathbf{r}^\alpha) Q_T(s,a).
\end{equation}

For a targeted attack, we define a \emph{target policy} as a set-value function $f(s)$ which maps each state to a set of target actions, i.e., $f : S\rightarrow 2^{A}$, where $f(s) \subseteq A$ for each state $s \in S$.
That is, the attacker aims to cause the learned policy $\tilde{\pi}$ to take one of a target actions $f(s)$, that is, $\tilde{\pi}(s;\mathbf{r}^\alpha) \subseteq f(s)$.
A natural special case is when $f(s)$ is a singleton for each state $s$.
This objective can be equivalently expressed in terms of the learned Q function $\tilde{Q}_T$ as the following condition in each state $s$:
\begin{equation}
\underset{a\in f(s)}{\max}\tilde{Q}_T(s,a;\mathbf{r}^\alpha) > \underset{a\notin f(s)}{\max}\tilde{Q}_T(s,a;\mathbf{r}^\alpha), \ \ \forall s : f(s) \ne \pi(s).
\end{equation}

We assume
that the current state $\sigma_t$ of the DQN algorithm is observed by the adversary at each time $t$.
This assumption amounts to the compromised scheduler being able to read memory.
This can be done either because the 
adversary has gained access to
kernel space, or through a side-channel attack that recovers memory contents~\cite{luo2020stealthy,ravensec_raid2019,kocher2019spectre}.

Our key observation is that whichever of the above goals the attacker chooses, since learning itself can be modeled as an MDP, the attacker's problem also becomes an MDP in which the state is
$o_t = (\sigma_t, \mathcal{D}_t)$, where $\mathcal{D}_t$ is the state of the attacker disk (i.e., rewards saved, and their current delay length).
The reward in this MDP is defined as above for both untargeted and targeted attacks.
Let $O$ denote the set of possible states in the attacker MDP.
Next, we define two types of reward-delay attacks, which then determine the action space.

\begin{figure}[h!]
  \centering
  \includegraphics[width=12.4cm, height=6.7cm]{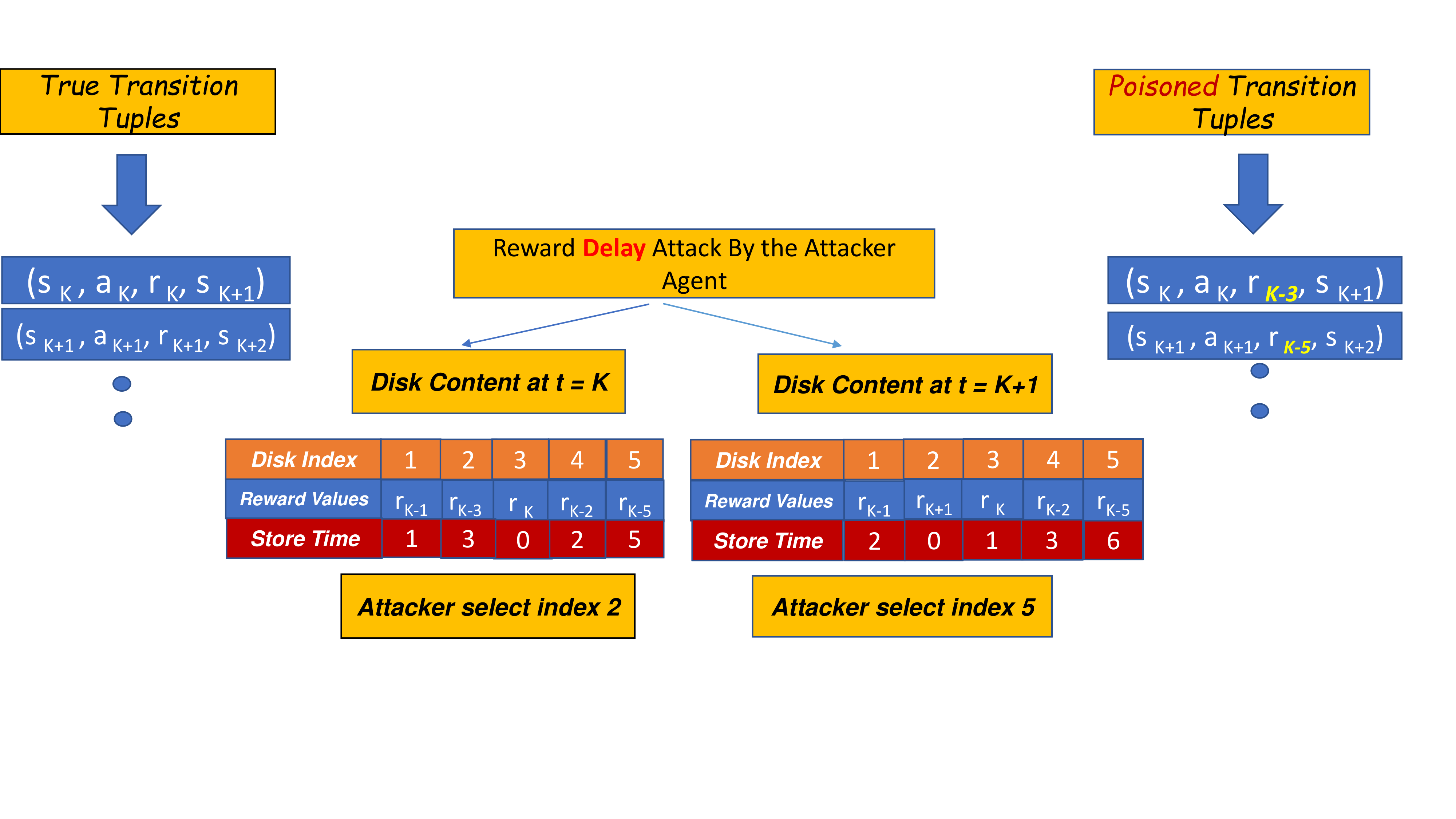}
  \caption{Reward Delay Attack Model.}
  \label{fig:reward_delay_attack_}
 \end{figure}

We consider two variants of the reward delay attack.
The first is a general \emph{reward delay} variant in which rewards can be swapped arbitrarily or dropped, with the only constraint that if a reward is delayed, it is by at most $\delta$ time steps.
The attacker also has the option of waiting at time t.
In particular, at time $t$, the attacker can publish (e.g., by prioritizing the scheduling of) a reward $r_t^\alpha \in \mathcal{D}_t$ selected from disk at time $t$, $\mathcal{D}_t$; the implied set of attacker actions at time $t$ is denoted by $A_t^\alpha = A^\alpha(o_t)$ (since it depends on the current state $o_t$).
As a result, the learner receives the transition tuple $(s_t,a_t,r_t^\alpha,s_{t+1})$ in place of $(s_t,a_t,r_t,s_{t+1})$, where the current reward $r_t$ is added to disk, obtaining the disk $\mathcal{D}_{t+1}$ for the next training step. We present an illustrative visualization of the \emph{reward delay} attack model in Figure~\ref{fig:reward_delay_attack_}.

\begin{figure}[h!]
  \centering
  \includegraphics[width=12.5cm, height=6.6cm]{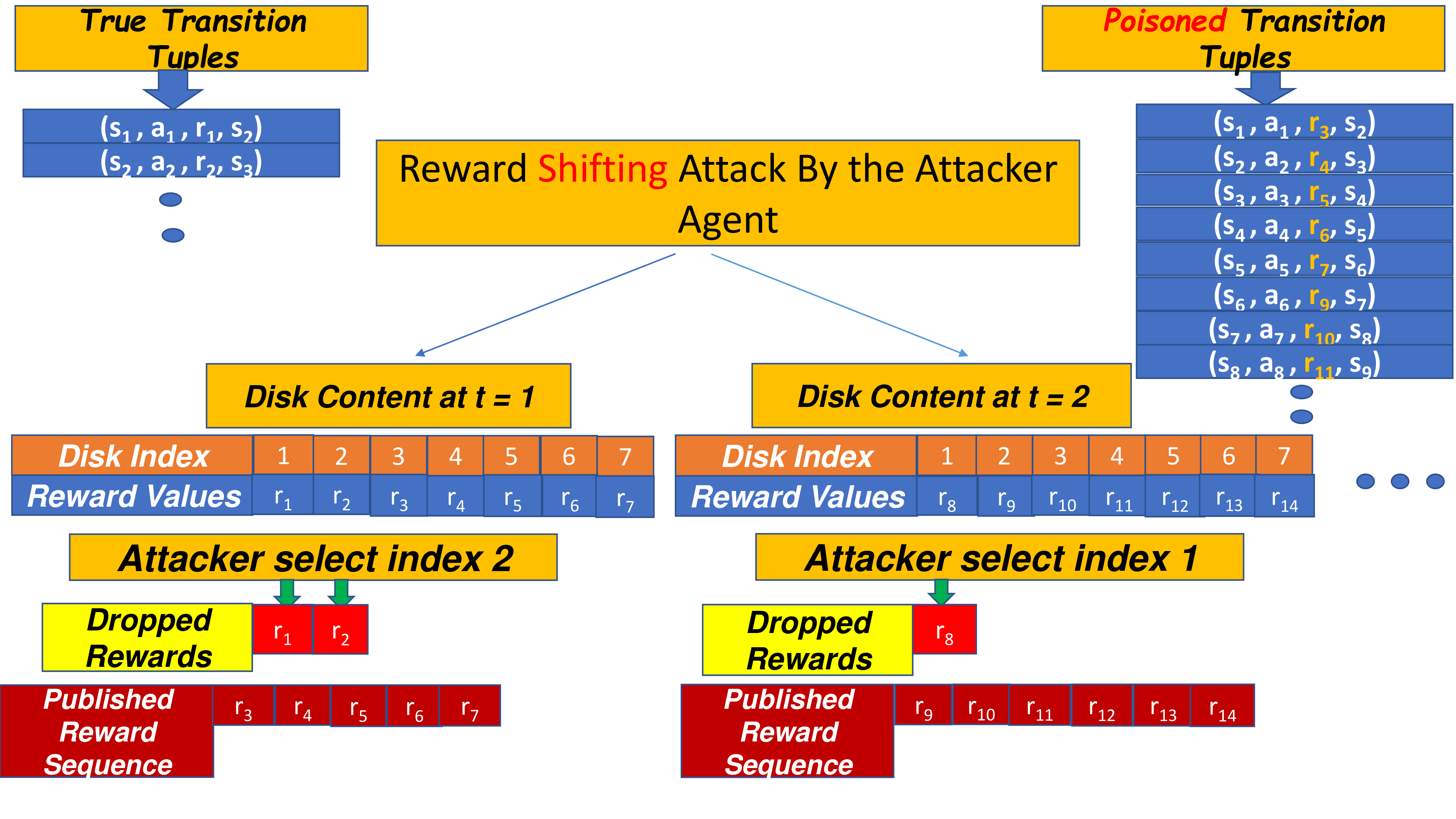}
  \caption{Reward Shifting Attack Model.}
  \label{fig:reward_shift_reward}
 \end{figure}

A minimal level of security that a learner can easily implement is to securely time stamp incoming sensor data.
Our goal is to evaluate how much impact this minimal level of protection has on the attack efficacy.
To this end, we introduce a second significantly more constrained attack variant that only allows \emph{reward shifting}: rewards can only be shifted forward (effectively, dropping some of these), but not arbitrarily swapped.
In reward shifting attacks, since the sequence must be preserved, any time the attacker selects a reward $r_t^\alpha \in \mathcal{D}_t$ to publish, it must be the case that the time stamp on this reward exceeds that of the reward published at time $t-1$.
Consequently, the disc $\mathcal{D}_{t+1}$ is updated with the actual reward $r_t$, but all rewards in $\mathcal{D}_t$ with time stamp earlier than $r_t^\alpha$ are also removed (effectively, dropped).
Additionally, the attacker has the option of waiting at time $t$, publishing a reward (or a sequence of rewards) at a later time point from the disk, which are then aligned in the corresponding temporal sequence with states and actions used for DQN updates. We present an illustrative visualization of the \emph{reward shifting} attack model in Figure~\ref{fig:reward_shift_reward}.
We denote the implied action space for the attacker in reward shifting attacks by $A_t^{\alpha,\text{shift}} = A^{\alpha,\text{shift}}(o_t)$.
Next, we present our algorithmic approaches for implementing the attack variants above.


\section{Algorithmic Approaches for Reward-Delay Attacks}
\label{sec:algo}
 \begin{figure*}[h!]
    \centering
    \begin{subfigure}[t]{0.50\textwidth}
        \centering
        \includegraphics[height=2.4in,width=2.33in]{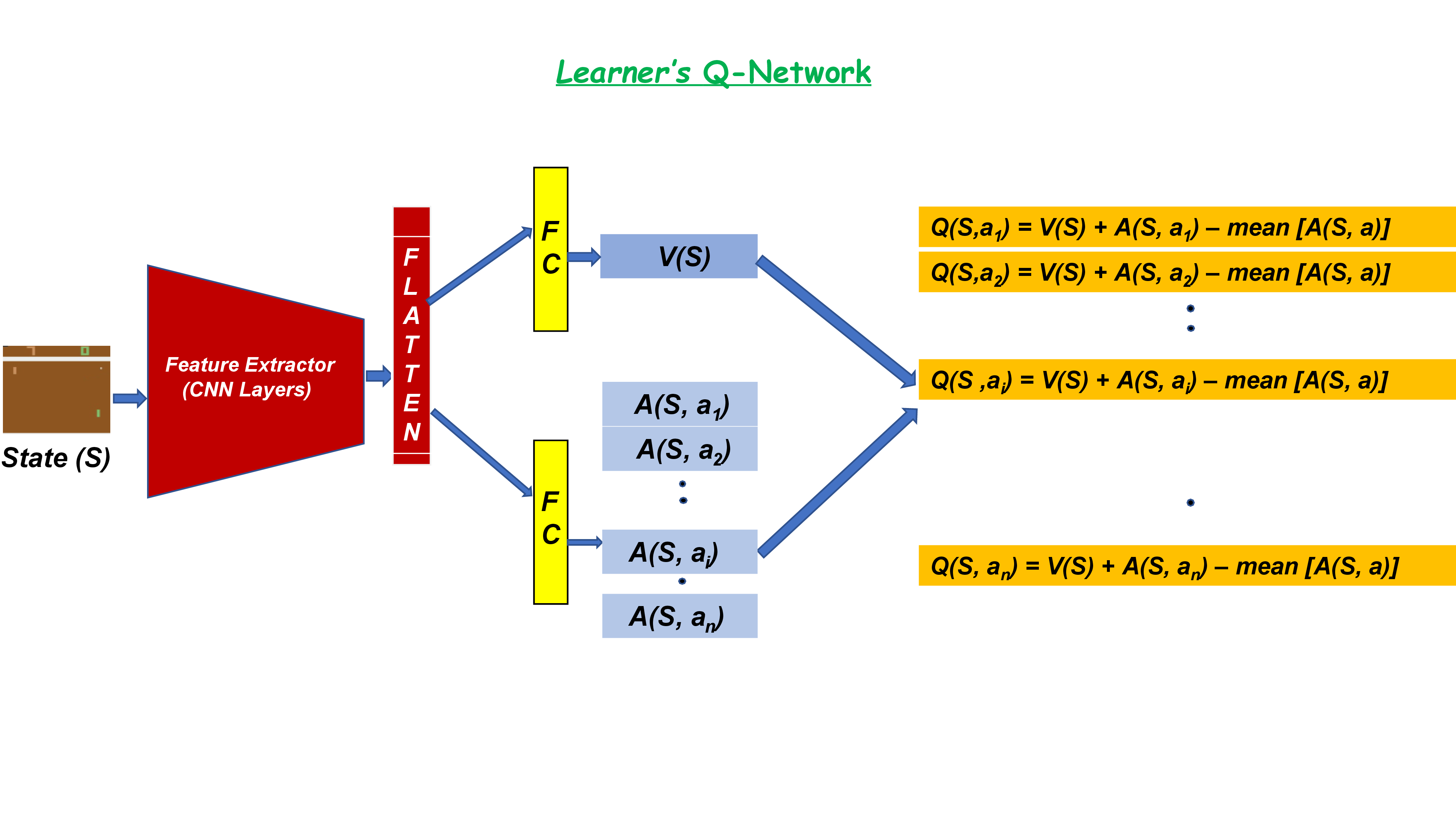}
    \end{subfigure}%
    ~ 
    \begin{subfigure}[t]{0.50\textwidth}
        \centering
        \includegraphics[height=2.4in,width=2.60in]{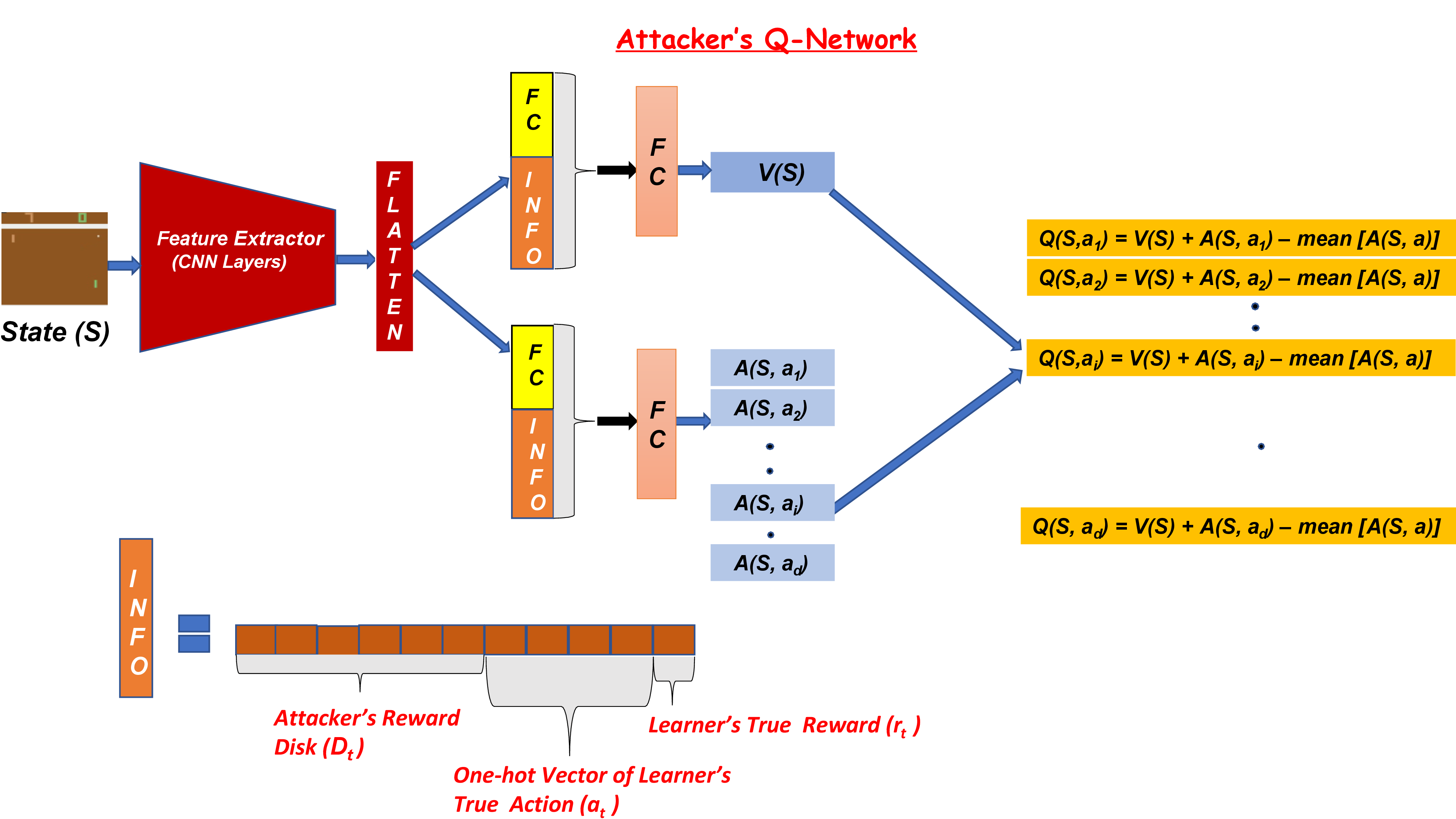}
    \end{subfigure}
    \caption{Learner's and Attacker's Q-network.}
\label{fig:architecture}
\end{figure*}
Recall that general reward-delay attacks can be represented by an attacker MDP with state space $O$ and action sets $A^\alpha(o)$ for the general reward-delay attacks, and $A^{\alpha,\text{shift}}(o)$ for reward shifting attacks, where $o \in O$ is the current state.
Since the state space is in general high-dimensional, it is natural to apply deep Q-learning attacks to learn an effective attack policy $\pi^\alpha(o)$.
In Figure~\ref{fig:architecture} we present both the learner's and attacker's Q-network architecture (as the latter is partly derived from the former).
An important practical challenge, however, is that delay reward signal so long considerably reduces efficacy of learning.
Consequently, our algorithmic approaches to the different types of attacks involve designing effective proxy-reward signals that can be computed in each time step $t$ of the learning process.

Since the reward shifting attack involves a considerably stronger constraint on what the attacker can do (which we model by modifying what information can be stored on disk $\mathcal{D}_t$ at time $t$ above), we further enhance our ability to effectively learn an attack policy in two ways.
First, we heavily leverage the \emph{wait} option by delaying attack choice until the disk $\mathcal{D}_t$ is full, in the sense that we can no longer wait without having to drop one of the rewards in the disk (which would otherwise exceed the delay time constraint $\delta$).
We then significantly simplify the attack strategy by  selecting a \emph{drop index} $i$ (where $0\leq i \leq K$) and dropping all the rewards in the disk $\mathcal{D}_t$ with index $\leq i$ from consideration.
Note that, $K < d$, where $d$ is the size of the attacker's disk.
The residual temporal sequence of rewards on disk is then published (as depicted in Figure~\ref{fig:reward_shift_reward}) and used to train the learner’s Q-network, and finally the attacker's disk is emptied.
This reduces the policy consideration space to only the choice of an index of the reward to drop given a full disk.

At this point, the only remaining piece of the attack approach is the design of the proxy reward function, which we turn to next.
Armed with appropriate proxy reward functions, we can apply any deep Q-network based algorithm for learning an attack policy for any of the attacks discussed above.

\subsection{Proxy Reward Design for Untargeted Attacks}

For untargeted attacks, suppose that $Q_t$ is the \emph{learner's} Q function in iteration $t$.
Since DQN updates are deterministic, the learner's Q function $Q_{t+1}$ can be precomputed for any reward $r_t^\alpha$ published by the attacker.
Let $s_t$ be the state observed by the learner at time $t$.
We propose the following proxy reward for the attacker which is used for the \emph{attacker's} DQN update:
\begin{equation}
\label{eq:Q-diff}
\begin{split}
\tilde{R}^{\text{att}}_{t} = \sum_{a \in A} - (\tilde{\pi}_{t+1}(s_t,a) \cdot Q_t(s_t,a)),\\
\text{where}\>\> \tilde{\pi}_{t+1}(s_t,a) =  \frac{\exp(\tilde{Q}_{t+1}(s_t, a))}{\sum_{a' \in A} \exp(\tilde{Q}_{t+1}(s_t, a'))}
\end{split}
\end{equation}
where $Q_t(s_t)$ is the \emph{true} Q function vector and $\tilde{Q}_{t+1}(s_t)$ is the \emph{proxy} Q function vector corresponding to all the learner's actions in step $t$ and $t+1$, respectively. Note that the learner's \emph{true} $Q_{t+1}(s_t)$ is obtained by updating $Q_t(s_t)$ using randomly sampled batch data stored in the learner's replay buffer, whereas, the learner's \emph{proxy} $\tilde{Q}_{t+1}(s_t)$ is obtained by updating $Q_t(s_t)$ using the recent transition tuples published by the attacker. 
The intuition for this proxy reward is that it accomplishes two things at once: first, by minimizing correlation between successive Q functions, the attacker minimizes the marginal impact of learning updates, thereby causing learning to fail, and second, if the learner happens to obtain a good estimate of the true Q function in iteration $t$, the quality of this function is actively reduced in iteration $t+1$.

\subsection{Proxy Reward Design for Targeted Attacks}

The intuition for our proposed proxy reward function in the case of targeted attacks is to maximize similarity between the target policy $f(s)$ and the policy induced by the current Q function $Q_t$.
However, since the policy induced by $Q_t$ is not differential, we replace it with a stochastic policy $\pi_t(s_t) = \text{softmax}({ Q_t(s_t)})$, where
\[
\pi_t(s_t,a) =  \frac{\exp(Q_t(s_t, a))}{\sum_{a' \in A} \exp(Q_t(s_t, a'))}.
\]
Further, we represent $f(s)$ as a vector $\hat{f}(s,a)$
\begin{equation}
    \hat{f}(s,a)\xleftarrow{}\left\{
    \begin{array}
    {r@{\quad\quad}l}
         \text{1,} & \text{if}\ a \in f(s,a) \\
         \text{0,}  & \text{o.w.}
    \end{array}
    \right.
\end{equation}
$\hat{f}(s)$ then denotes the binary vector corresponding to $f(s)$.
We then define the proxy reward for a targeted attack as follows:
\begin{align}
\label{eq:targeted}
\tilde{R}^{\text{att}}_{t} = \mathrm{sign}\{\mathcal{L}_{CE}(\pi_t(s_t), \hat{f}(s_t)) - \mathcal{L}_{CE}(\tilde{\pi}_{t+1}(s_t), \hat{f}(s_t))\},
\end{align}
where $\mathcal{L}_{CE}$ is the cross-entropy loss. The reward function in \eqref{eq:targeted} suggests that, an attacker receives a positive reward, only if the performed actions of the attacker (i.e. choosing $r^{\alpha}_{t}$ from the disk $\mathcal{D}_{t}$ ) steers the learner's updated proxy Q-value distribution $\tilde{\pi}_{t+1}(s_t)$ to be more aligned with the target Q-table distribution corresponds to target policy $f(s)$ compared to the learner's previous true Q-value distribution $\pi_{t}(s_t)$.

\subsection{Rule Based Targeted Reward Delay Attack} In addition to a targeted reward delay attack strategy that requires a proxy reward computation of the attacker as defined in equation \eqref{eq:targeted}, we also propose a simple rule based strategy in order to feed the attacker a reward at a given state as defined in equation \eqref{eq:rule_}. Note that the attacker's disk configuration at current time step (t) is represented as $\mathcal{D}_{t}$. According to \eqref{eq:rule_}, the attacker feeds back a high reward to the learner, if the learner acts in a way that is preferred by the attacker at any given target state.

\begin{equation}
        \tilde{R}^{\text{att}}_{t}\xleftarrow{}\left\{
        \begin{array}
        {r@{\quad\quad}l}
            \text{\emph{Maximum} Reward in $\mathcal{D}_{t}$,} & \text{if}\ \arg\max_{a} (Q_{t}(S_t, a)) \in a^{T} \> \text{and}\>  S_t \in S^{\prime} \\
            \text{\emph{Minimum} Reward in $\mathcal{D}_{t}$,} & \text{if}\
            \arg\max_{a} (Q_{t}(S_t, a)) \notin a^{T} \> \text{and}\> S_t \in S^{\prime} \\
            \text{\emph{Random} Reward in $\mathcal{D}_{t}$,} & \text{if}\
            S_t \notin S^{\prime} \\
        \end{array}
        \right. 
        \label{eq:rule_}
\end{equation}

\section{Results}


We evaluate the effectiveness of the proposed attack approaches 
on the Pong and Breakout Atari-2600 environments in OpenAI Gym~\cite{brockman2016openai} on top of the Arcade Learning Environment~\cite{bellemare2013arcade}.
The states in those environments are high dimensional RGB images with dimensions (210 * 160 * 3) and discrete actions that control the agent to accomplish certain tasks. Specifically, we leverage the \emph{NoFrameskip-v4} version for all our experiments, where the randomness that influences the environment dynamics can be fully controlled by fixing the random seed of the environment at the beginning of an episode. Please note that we used a standard computing server with 3 GeForce GTX 1080 Ti GPUs each of 12GB for all the experiments in our work. We choose the Double DQN algorithm~\cite{van2016deep} with the Duelling style architecture~\cite{wang2016dueling} as the reference Q network for both the learner and attacker agents. Note that our proposed reward-delay attack and reward-shifting attack strategies can be easily applied to other DQN based learning algorithms without requiring further modifications. 
Unless noted otherwise, we set $\delta$ (i.e. the maximum number of time steps a reward can be delayed) to be 8. 
In the case of a \emph{reward shifting} attack, we choose the maximum value of \enquote{drop-index}($K$) to be 4, and attacker's maximum disk size (effectively, maximum wait time before implementing the attack) to be 8.
We also perform experiments to show the impact of the reward delay attack for different choices of attack hyper-parameters.

We compare our approaches to two baselines: \emph{random attack} and \emph{fixed-delay attack}. 
In the random attack, an attacker chooses to publish a reward randomly from the disk at every time-step. 
In the fixed-delay attack, an attacker delays the reward signal by $\delta$ time steps.
In addition, in the reward shifting attack setting, we use a \emph{random reward Shift} baseline, where in place of an attacker agent, we randomly select a value for \enquote{drop-index} (ranging from index 0 to $K$) to drop the reward. Apart from this step, the random baseline attack operates exactly as the \emph{reward shifting} attack. 

For untargeted attacks, our measure of effectiveness is the expected total reward at test time.
In the case of targeted attacks, we evaluate the \emph{success rate} of the attacks, measured as follows:
\begin{equation}
\label{eq:success_rate}
Success Rate(SR) = \frac{\sum_{s\in S^{\prime}} \mathbb{I}[a_{t} \in f(s)]}{\textit{No. of times agent visits target states}},
\end{equation}
where $\mathbb{I}[.]$ is an indicator function and $S'$ is a set of \emph{target states} in which the attacker has a non-trivial preference over which action is played.
Equation~\eqref{eq:success_rate} yields the fraction of times the policy learned by the targeted RL agent chooses an action in the attacker's target set $f(s)$ in target states $s \in S'$.

In order to generate target policies $f(s)$, and in particular which states constitute target states $S'$,
we leverage an (approximately) optimal Q-network (learned without attacks), denoted by $Q^{*}$, which gives us a way to decide a subset of states that we target given the target action sets preferred by the attacker; for the remaining states, $f(s)$ allows any action, that is, the attacker is indifferent.
Next, we define a state-independent set of target actions $a^T$; these will be target actions for a subset of target states, which we choose dynamically using the following rule:
\begin{equation}
        s_t\xleftarrow{}\left\{
        \begin{array}
        {r@{\quad\quad}l}
            \text{Is a Target State,} & \text{if}\ \arg\max_{a} (Q^{\textbf{*}}(S_t, a)) \notin a^{T}.\\
            \text{Not a Target State,} & \text{otherwise.}
        \end{array}
        \right. 
        \label{eq:target_state_}
\end{equation}
We describe our choice of the set $a^T$ in the concrete experiment domains below. Note that at every time step in an episode, the learner interacts with the environment, but instead of receiving the true reward from the environment, the learner receives a poisoned reward published by the attacker. We train the learner's Q network with the modified reward sequence published by the attacker and also update the learner's Q network parameter following the Double DQN update rule. In parallel, we train and update the attacker's Q network following the Double DQN update rule as described above (see also Algorithms 1 and 2 in the Supplement).
After the completion of each episode, we evaluate the learner's performance on test episodes.
We report our results (cumulative rewards, etc) obtained on these test episodes.

\subsection{Untargeted Reward-Delay Attacks}

\begin{figure*}[h]
    \centering
    \begin{subfigure}[t]{0.50\textwidth}
        \centering
        \includegraphics[height=1.8in,width=2.63in]{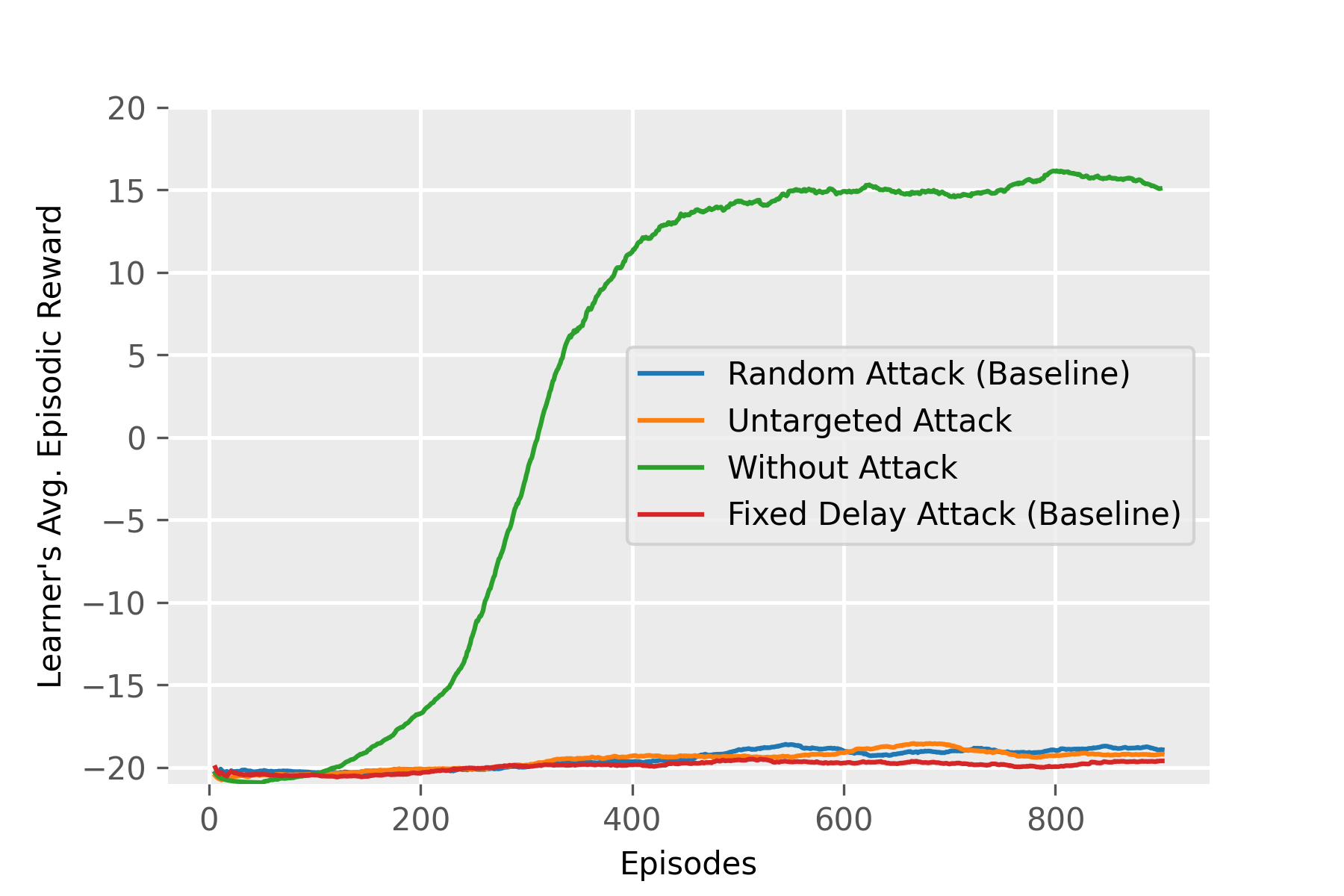}
        \caption{Reward Comparison on Atari Pong.}
    \end{subfigure}%
    ~ 
    \begin{subfigure}[t]{0.50\textwidth}
        \centering
        \includegraphics[height=1.8in,width=2.63in]{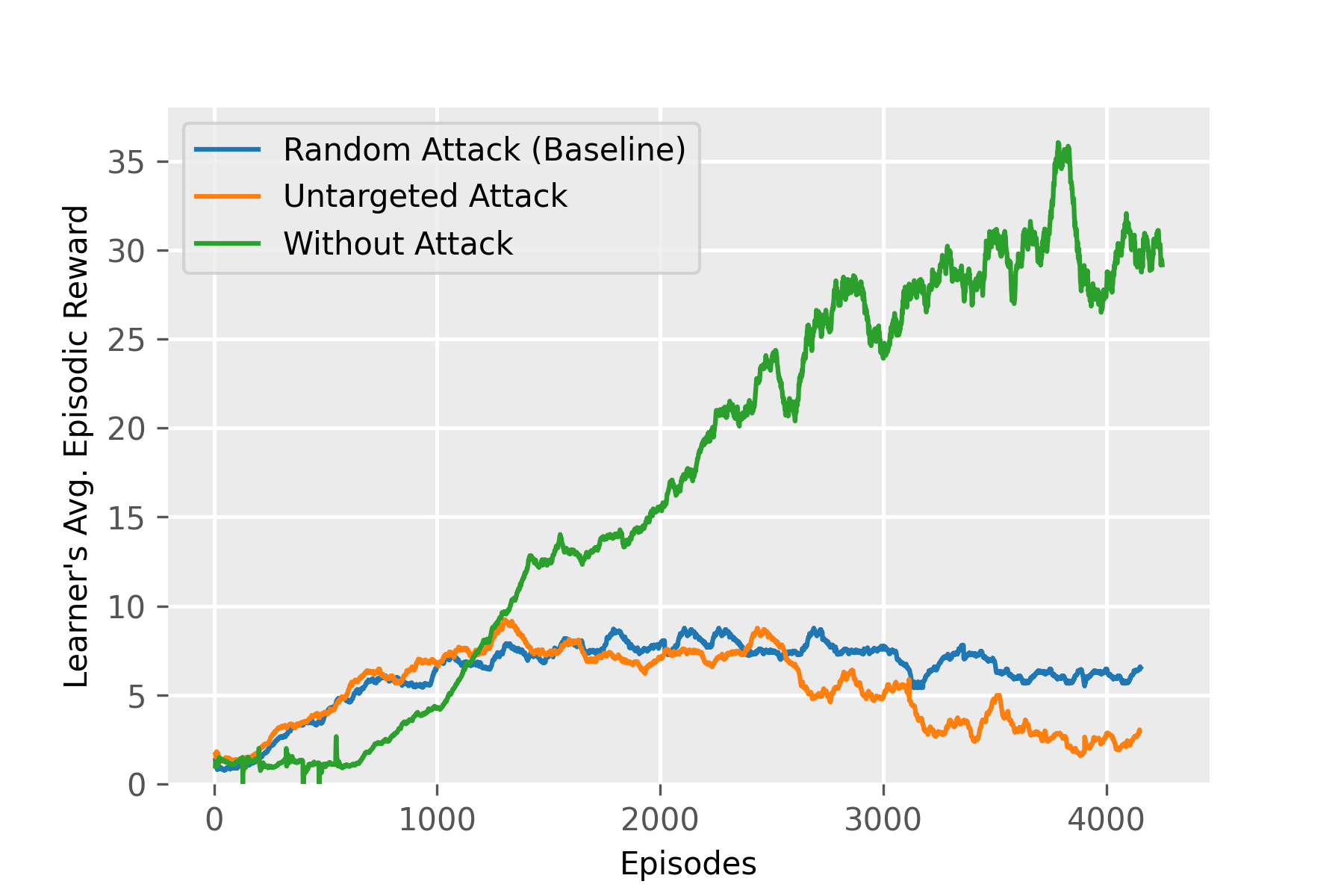}
        \caption{Reward Comparison on Atari Breakout.}
    \end{subfigure}
    \caption{Effect of Reward Delay Attack on Minimizing Reward.}
    \label{fig:reward_delay_reward}
\end{figure*}

We begin by evaluating the efficacy of the proposed approaches for accomplishing the goals of untargeted attacks.
Figure \ref{fig:reward_delay_reward}
presents the total reward obtained during evaluation as a function of the number of training episodes in Atari Pong and Breakout, respectively.
We observe that essentially all baselines perform nearly the same as each other and as our attack, with reward nearly zero in the case of Atari Breakout and nearly -21 in the case of Atari Pong.
This clearly contrasts with normal training, which is highly effective.
Consequently, in the untargeted setting, even if we limit considerably by how much rewards can be delayed, essentially any reshuffling of rewards entirely prevents effective learning.



\begin{figure*}[h!]
    \centering
    \begin{subfigure}[t]{0.50\textwidth}
        \centering
        \includegraphics[height=1.8in,width=2.63in]{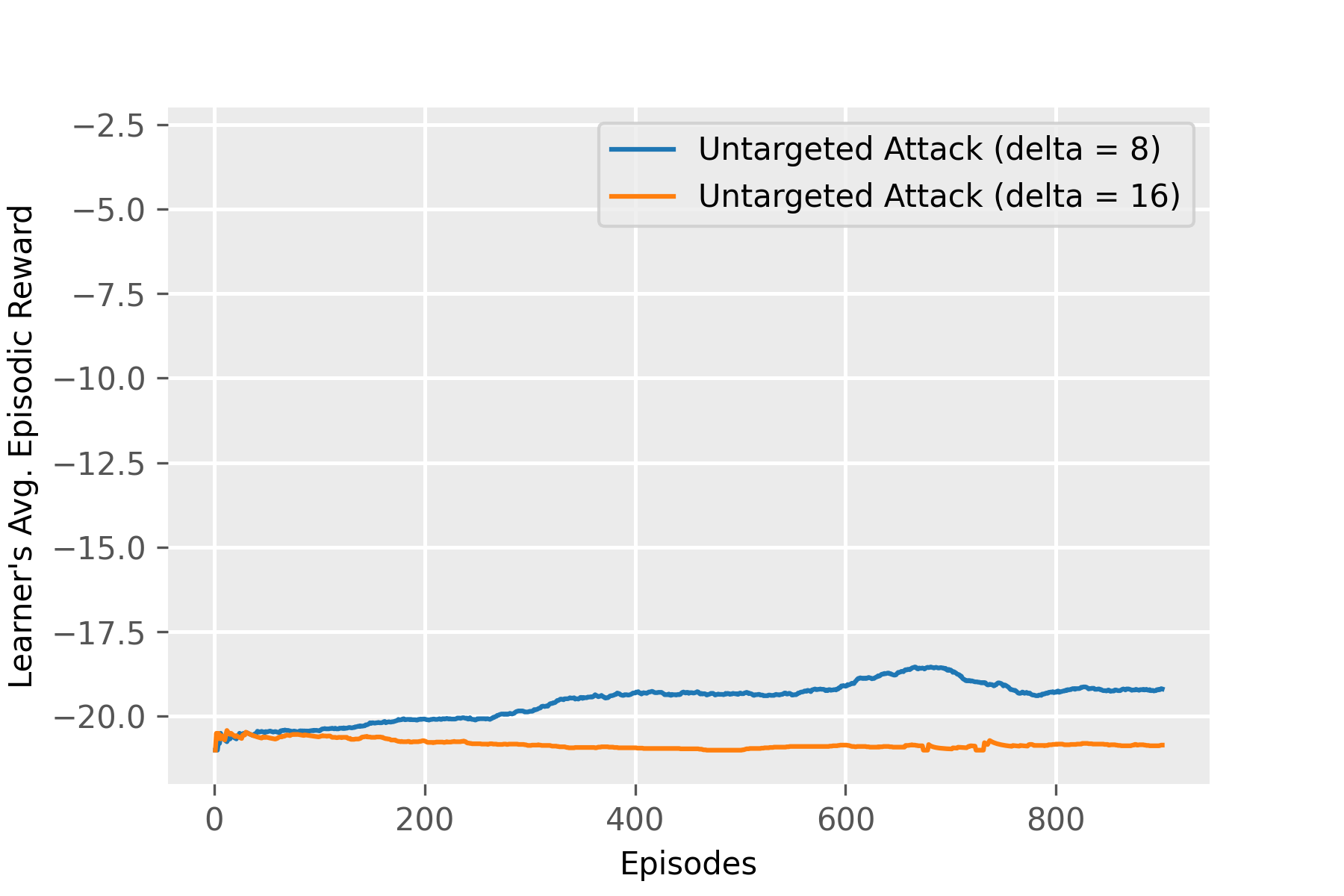}
        \caption{Effect of $\delta$ on Untargeted Reward Delay Attack Strategy on Atari Pong environment.}
    \end{subfigure}%
    ~ 
    \begin{subfigure}[t]{0.50\textwidth}
        \centering
        \includegraphics[height=1.8in,width=2.63in]{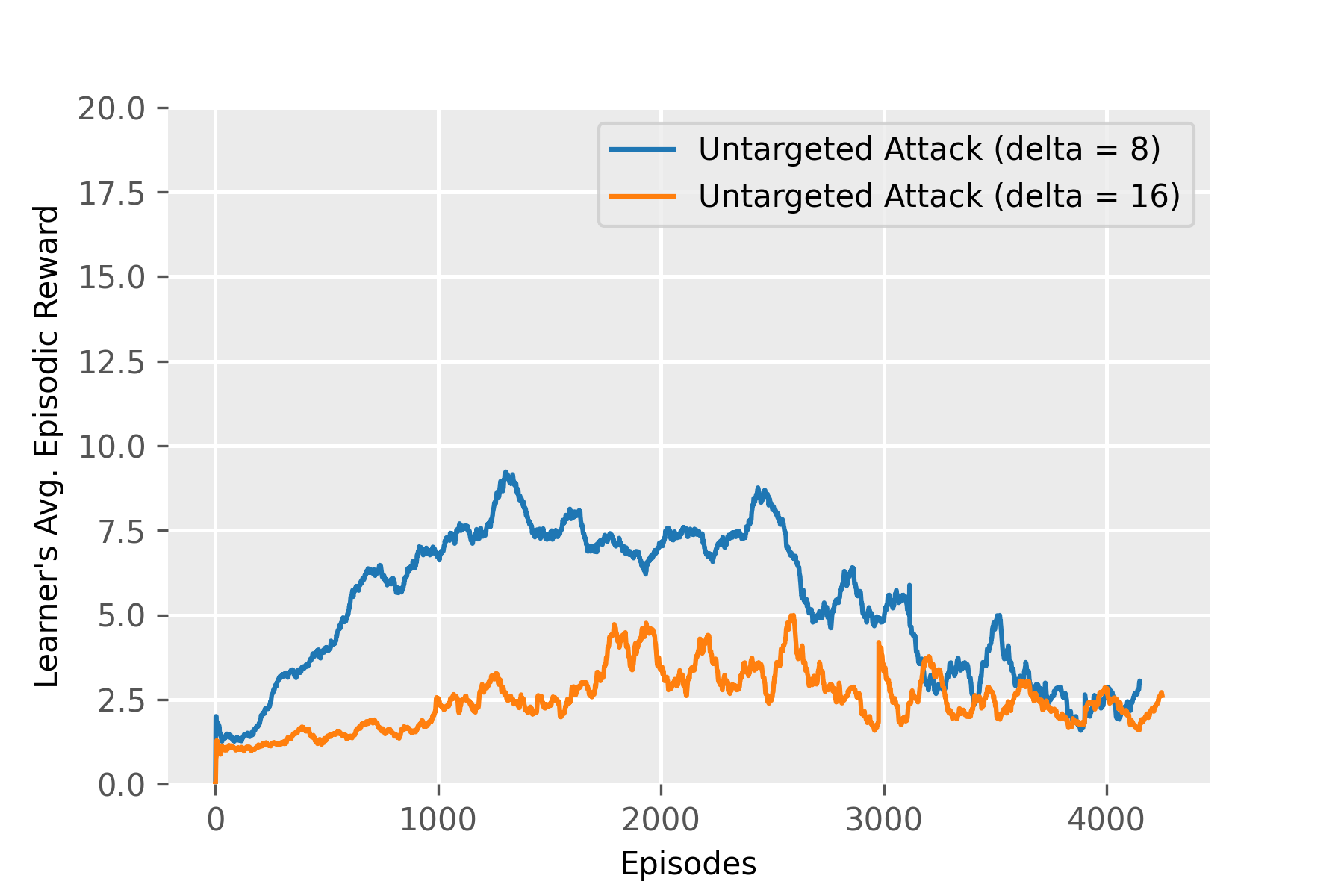}
        \caption{Effect of $\delta$ on Untargeted Reward Delay Attack Strategy on Breakout environment.}
    \end{subfigure}
    \caption{Effect of $\delta$.}
    \label{fig:effect_delta}
\end{figure*}

We further investigate the efficacy of the untargeted reward delay attacks as we change $\delta$, the maximum delay we can add to a reward (i.e., the maximum we can shift reward back in time relative to the rest of DQN update information), from 8 (the default value in experiments) to 16.
As Figure \ref{fig:effect_delta} shows, we see an improvement in the attack efficacy as would be expected intuitively; what is surprising, however, is that this improvement is extremely slight, even though we doubled the amount of time the reward can be delayed.
Our results thus suggest that even a relatively short delay in the reward signal can lead DQN learning to be entirely ineffective.

\begin{figure*}[h]
    \centering
    \begin{subfigure}[t]{0.50\textwidth}
        \centering
        \includegraphics[height=1.8in,width=2.63in]{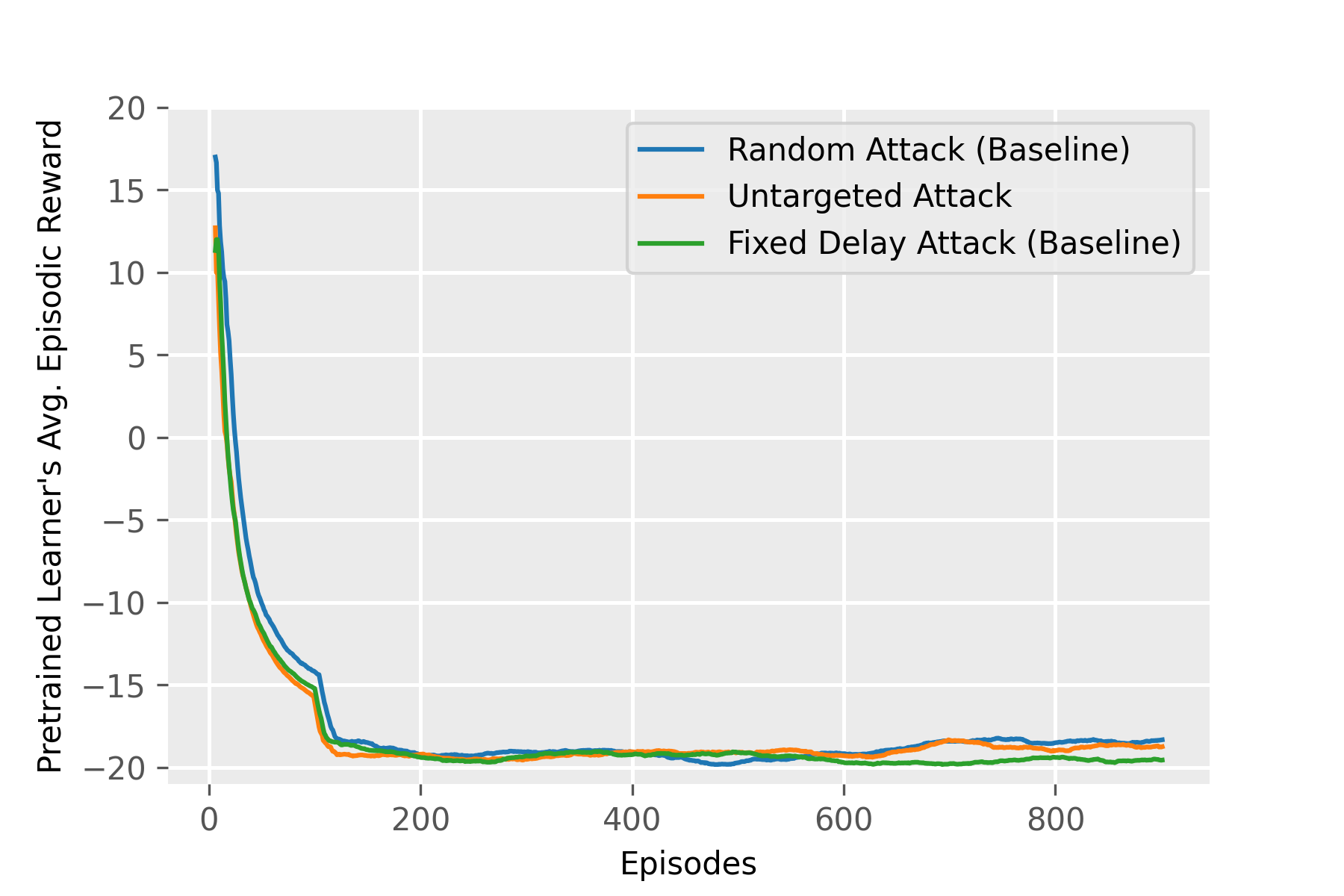}
        \caption{Effect of Untargeted Reward Delay Attack Strategy on a Pre-trained Q network on Atari Pong environment.}
    \end{subfigure}%
    ~ 
    \begin{subfigure}[t]{0.50\textwidth}
        \centering
        \includegraphics[height=1.8in,width=2.63in]{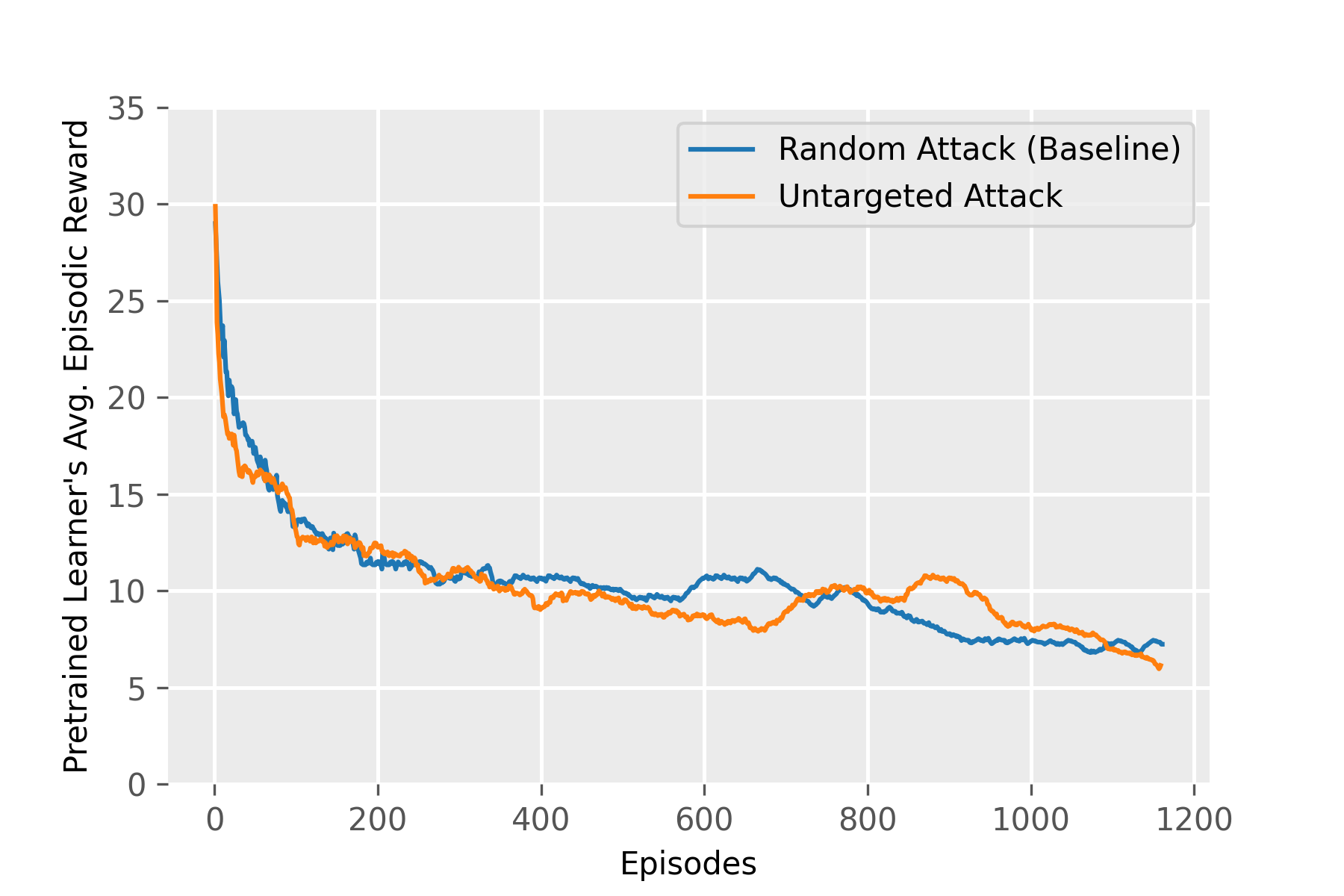}
        \caption{Effect of Untargeted Reward Delay Attack Strategy on a Pre-trained Q network on Breakout environment.}
    \end{subfigure}
    \caption{Effect of Reward Delay Attack on a Pre-trained Q network.}
    \label{fig:effect_untarget_pretrain}
\end{figure*}

We also examine the effectiveness of the untargeted reward delay attacks on a pre-trained Q network. Figure \ref{fig:effect_untarget_pretrain} depicts the net reward obtained by the pre-trained Q network during evaluation as a function of the number of training episodes in Atari Pong and Breakout, respectively. Our experimental finding indicates that, the reward decays exponentially as training progresses. We also notice that even naive baseline untargeted attacks are as effective as the untargeted reward delay attack in reducing the reward of the pretrained network. Such observations shows the importance of reward synchrony at every phase of deep reinforcement learning training.

Next, we evaluate the efficacy of our approach in the context of the far more challenging targeted attacks.





\subsection{Targeted Reward-Delay Attacks}

In targeted attacks, we aim to achieve a particular target action (or one of a set of actions) in a subset of \emph{target states}, with the attacker indifferent about which action is taken in the remaining states.
In both Pong and Breakout environments, we chose \emph{do not move} as the target action for this evaluation.
Target states were defined using the condition in Equation \eqref{eq:target_state_}.

In Figure \ref{fig:reward_delay_SR}, we compare the efficacy of our approach for targeted attacks compared to our baseline approach.
Here we can see that the proposed targeted attack approaches are considerably more effective than the baseline, with success rate significantly higher than the best baseline in both Pong and Breakout.
Interestingly, we can also see that while the proposed attack improves in efficacy with the number of training episodes, the baselines either have a constant success rate (essentially due entirely to chance), or the success rate of these may even decrease (we can see a mild decrease in the case of Pong, in particular). We also observe that the rule based reward delay targeted attack strategy is highly effective in achieving the targeted attack objective. \\

\begin{figure*}[h]
    \centering
    \begin{subfigure}[t]{0.50\textwidth}
        \centering
        \includegraphics[height=1.7in,width=2.63in]{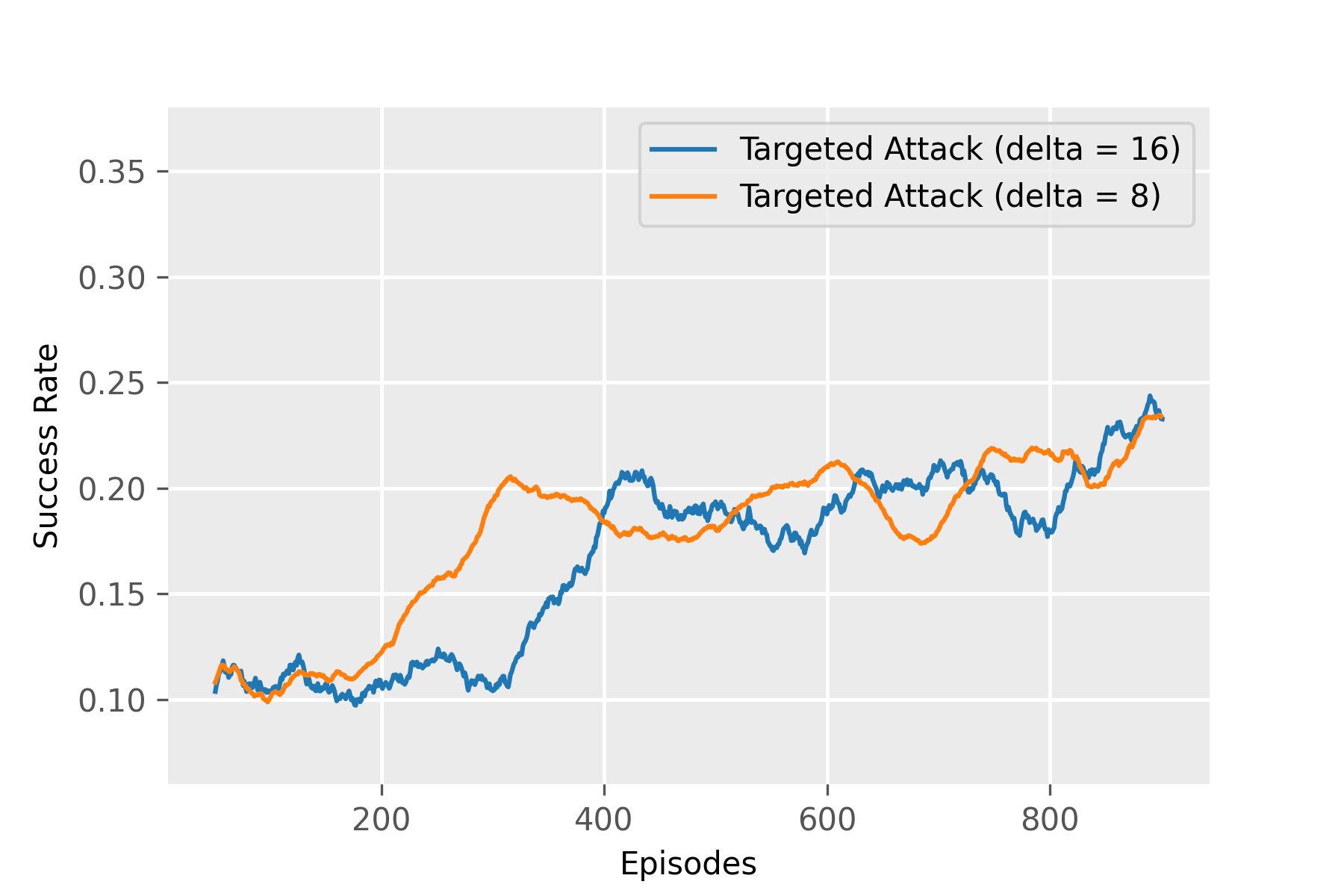}
        \caption{Effect of $\delta$ on Targeted Reward Delay Attack Strategy on Atari Pong environment.}
    \end{subfigure}%
    ~ 
    \begin{subfigure}[t]{0.50\textwidth}
        \centering
        \includegraphics[height=1.7in,width=2.63in]{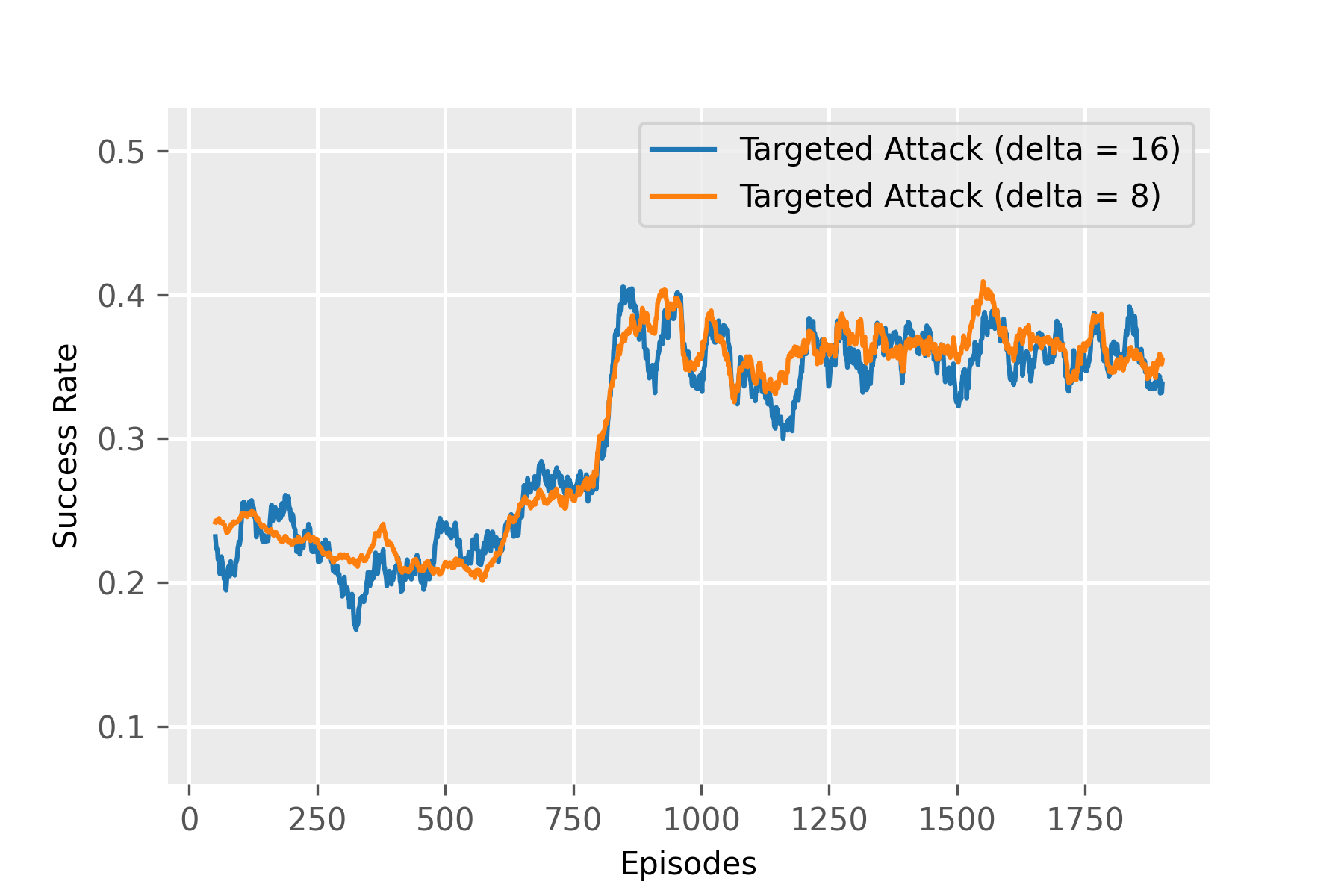}
        \caption{Effect of $\delta$ on Targeted Reward Delay Attack Strategy on Breakout environment.}
    \end{subfigure}
    \caption{Effect of $\delta$.}
    \label{fig:effect_delta_target}
\end{figure*}

We further look into the effectiveness of targeted reward delay attacks as we vary $\delta$, the maximum delay we can add to a reward. We present those results in figure \ref{fig:effect_delta_target}. We observe that the \emph{success rate} stays the same as we change the $\delta$ in both Atari Pong and Breakout environments.

Next, we assess the effectiveness of reward-shifting attacks, which are more constrained than reward delay attacks.

\begin{figure*}[h!]
    \centering
    \begin{subfigure}[t]{0.50\textwidth}
        \centering
        \includegraphics[height=1.7in,width=2.63in]{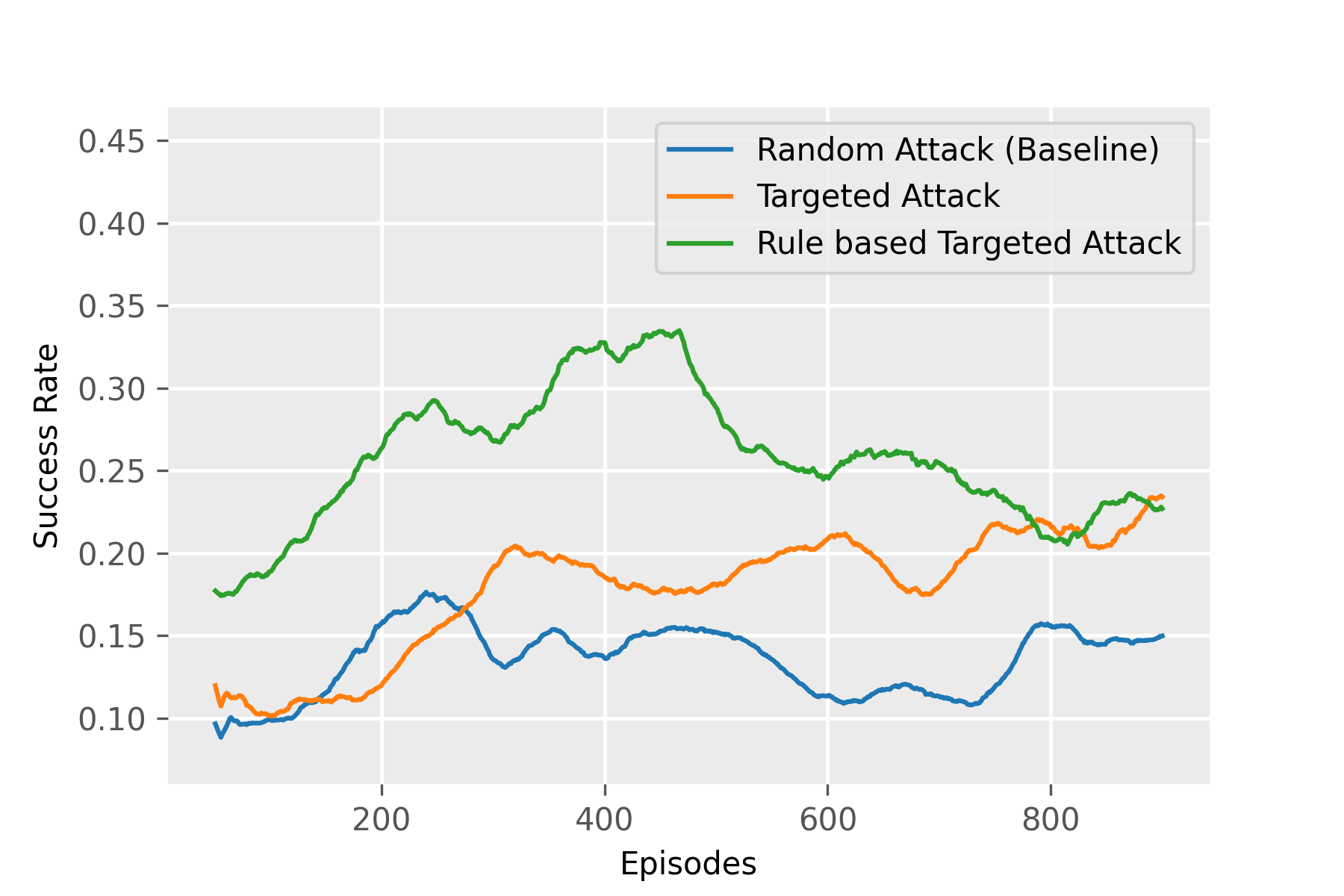}
        \caption{Success Rate comparison on Atari Pong.}
    \end{subfigure}%
    ~ 
    \begin{subfigure}[t]{0.50\textwidth}
        \centering
        \includegraphics[height=1.7in,width=2.63in]{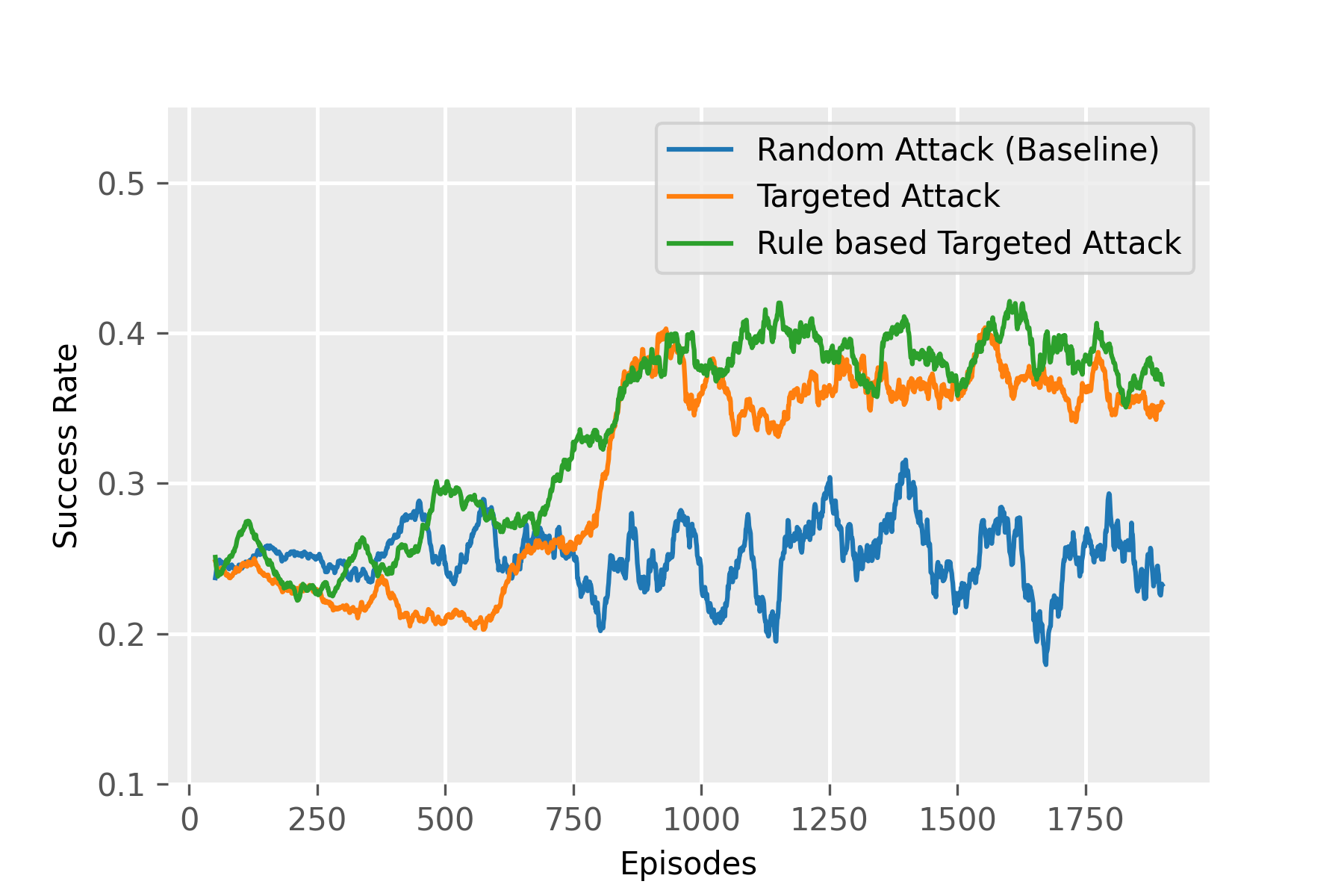}
        \caption{Success Rate comparison on Atari Breakout.}
    \end{subfigure}
    \label{fig:reward_delay_SR}
    \caption{Success Rate Comparison of Reward Delay Attack.}
    \label{fig:reward_delay_SR}
\end{figure*}
\vspace{-6pt}
\vspace{-9pt}

\subsection{Reward Shifting Attacks}
\vspace{-2pt}
We now turn to evaluating the effectiveness of a simple defense in which we ensure that rewards cannot be shuffled out of sequence.
To this end, we evaluate the efficacy of the proposed reward shifting attacks, and compare that to our observations of the efficacy of reward delay attacks above.

\begin{figure*}[h!]
    \centering
    \begin{subfigure}[t]{0.50\textwidth}
        \centering
        \includegraphics[height=1.7in,width=2.63in]{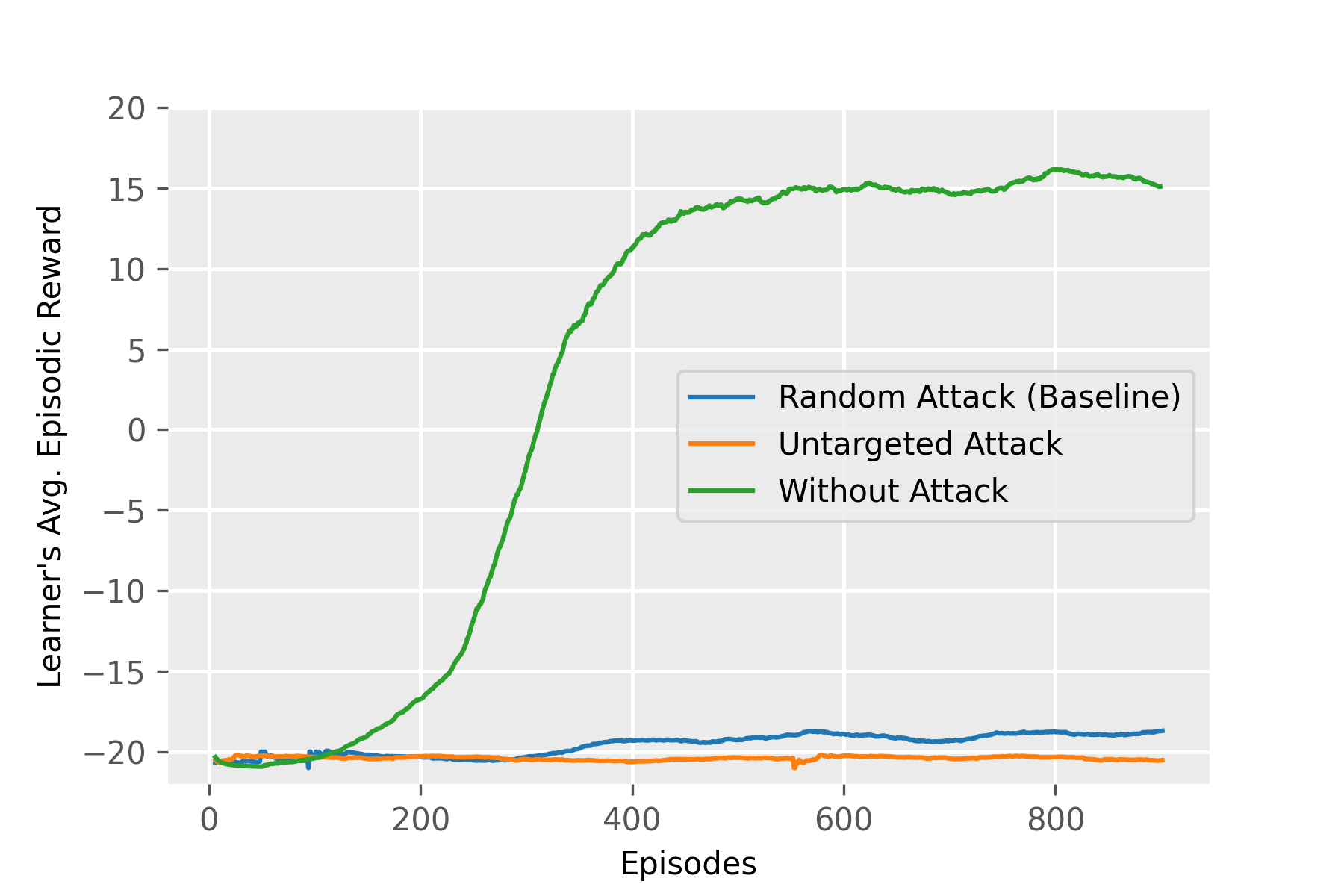}
        \caption{Reward Comparison on Atari Pong.}
    \end{subfigure}%
    ~ 
    \begin{subfigure}[t]{0.50\textwidth}
        \centering
        \includegraphics[height=1.7in,width=2.63in]{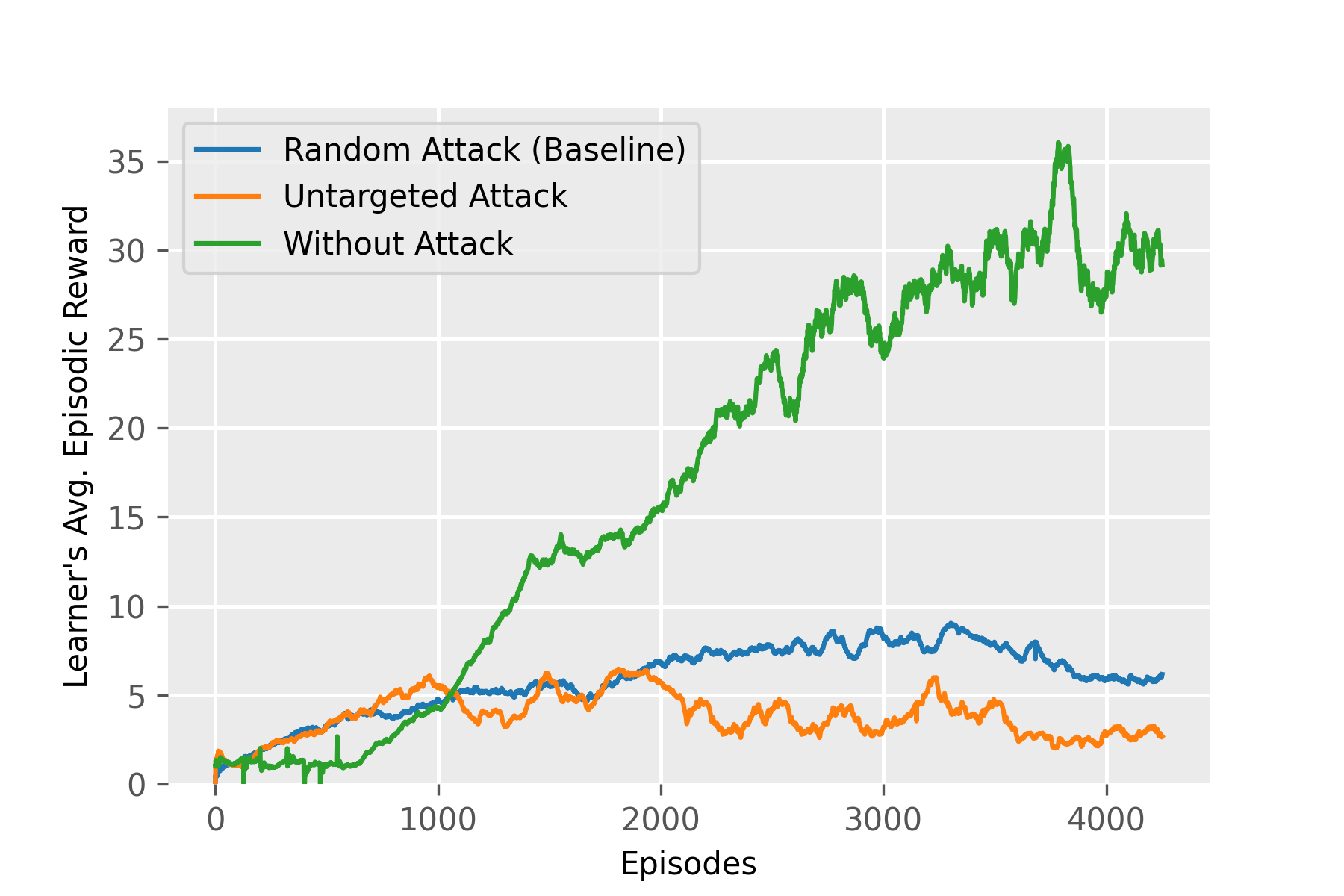}
        \caption{Reward Comparison on Atari Breakout.}
    \end{subfigure}
    \caption{Effect of Reward Shifting Attack on Minimizing Reward.}
    \label{fig:drop_shift_reward}
\end{figure*}

\begin{figure*}[h!]
    \begin{subfigure}[t]{0.50\textwidth}
        \centering
        \includegraphics[height=1.55in,width = 2.63in]{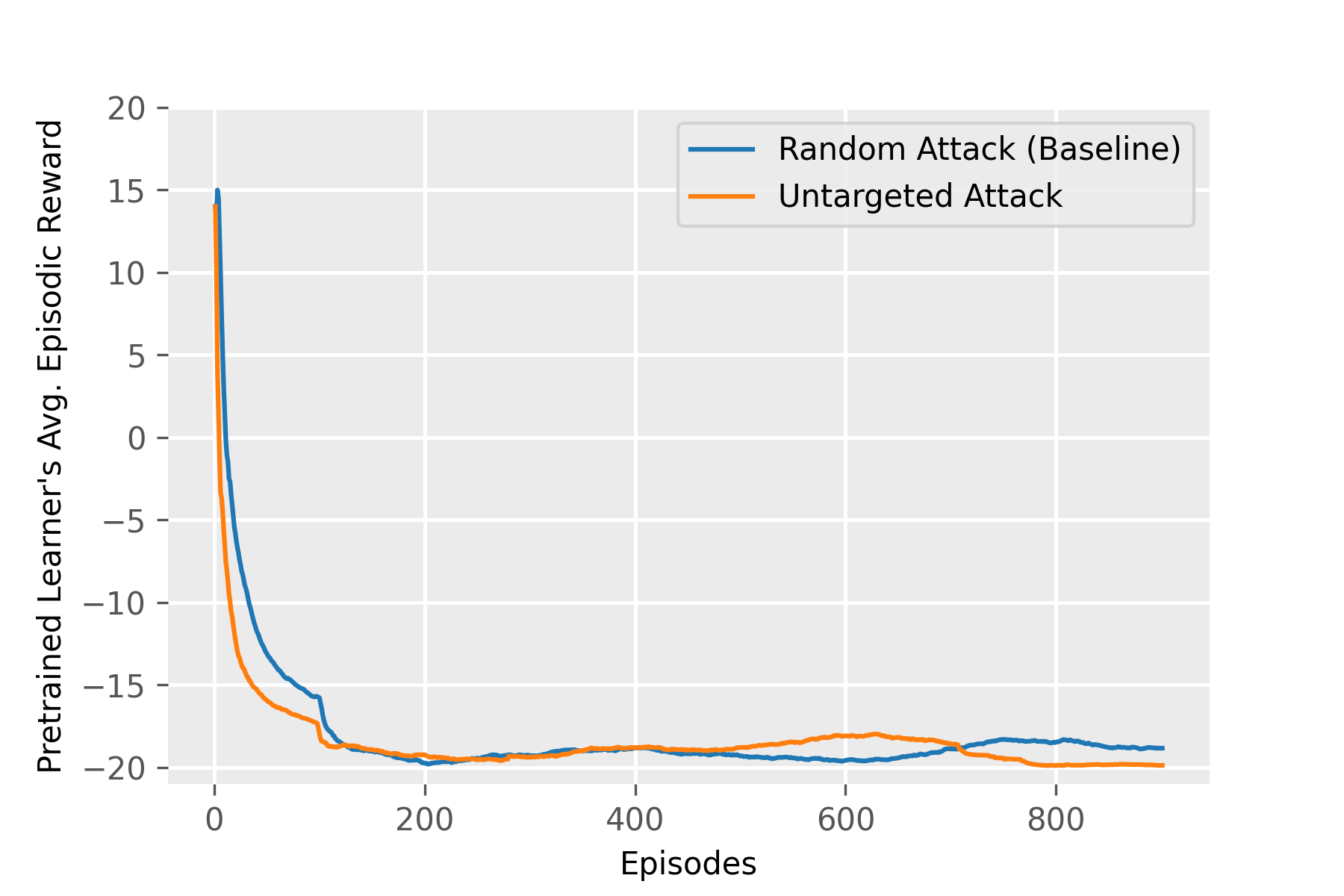}
        \caption{Effect of Untargeted Reward Shifting Attack on a Pre-trained Q network on Pong environment.}
    \end{subfigure}%
    ~ 
    \begin{subfigure}[t]{0.50\textwidth}
        \includegraphics[height=1.55in, width=2.63in]{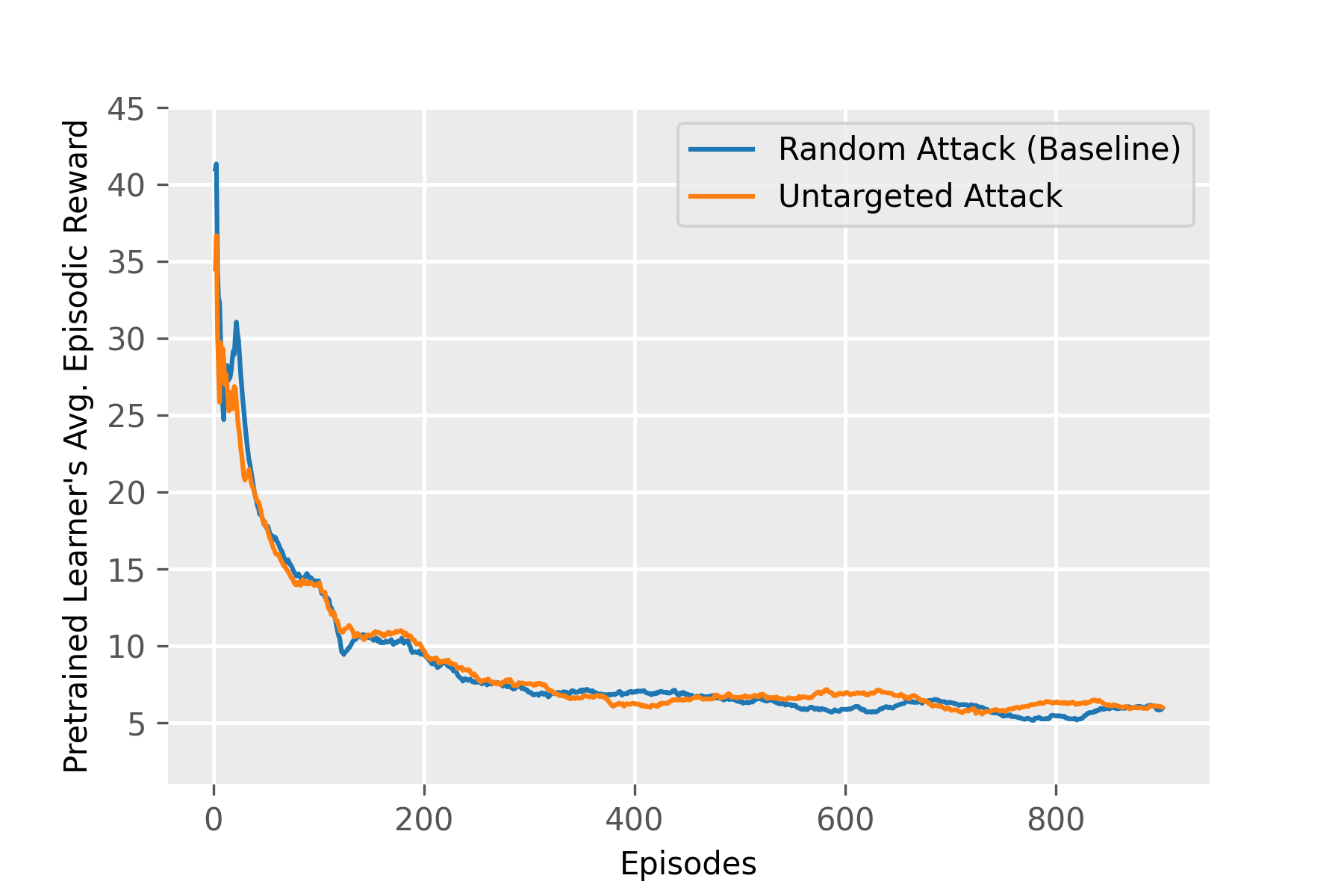}
        \caption{Effect of Untargeted Reward Shifting Attack on a Pre-trained Q network on Breakout env.}
    \end{subfigure}
    \caption{Effect of Reward Shifting Attack on a Pre-trained Q network.}
    \label{fig:drop_shift_trained}
\end{figure*}

In Figure \ref{fig:drop_shift_reward}, we present the results of the \emph{reward shifting} attack in the Pong and Breakout environments. 
First, we observe that both the baseline and our untargeted reward shifting attacks are as effective as any of the attacks without the sequence-preserving constraint.  
Moreover, the proposed untargeted attack is now tangibly better than its baseline (random) counterpart, with the gap increasing with the number of episodes. In addition, Figure~\ref{fig:drop_shift_trained} shows that the net episodic reward of a pre-trained Q network drops exponentially when trained with the untargeted reward shifting attack strategy. We also found that even a random reward shifting attack is highly capable of reducing the reward of a pre-trained Q network. Such findings indicate the adverse impact of incorrect reward timing on deep reinforcement learning. So, not only the \emph{ordering of the reward sequence}, but the \emph{precise timings of the reward sequence} are also very important for efficient deep reinforcement learning.

Nevertheless, the defense is clearly not effective in mitigating the untargeted reward shifting attacks.
In the Pong environment, our proposed untargeted reward shifting attack yields reward near -20---that is almost what was achieved without any mitigation at all, but a far cry from the result of nominal training.
Similarly, the reward after our attack in the Breakout environment is still far below what is achievable without attack.

\begin{figure*}[h!]
    \centering
    \begin{subfigure}[t]{0.50\textwidth}
        \centering
        \includegraphics[height=1.55in,width=2.63in]{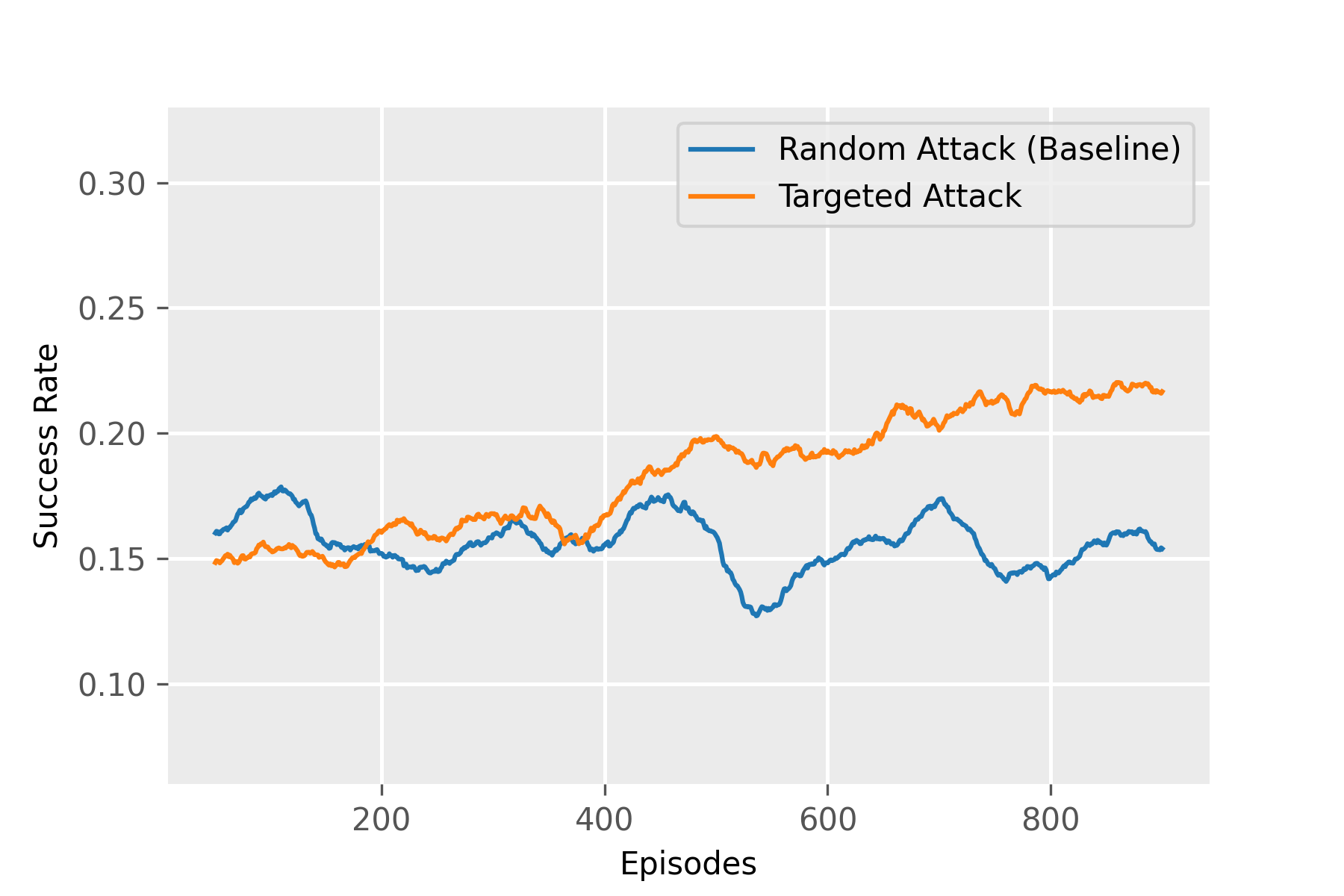}
        \caption{Success Rate Comparison on Atari Pong.}
    \end{subfigure}%
    ~ 
    \begin{subfigure}[t]{0.50\textwidth}
        \centering
        \includegraphics[height=1.55in,width=2.63in]{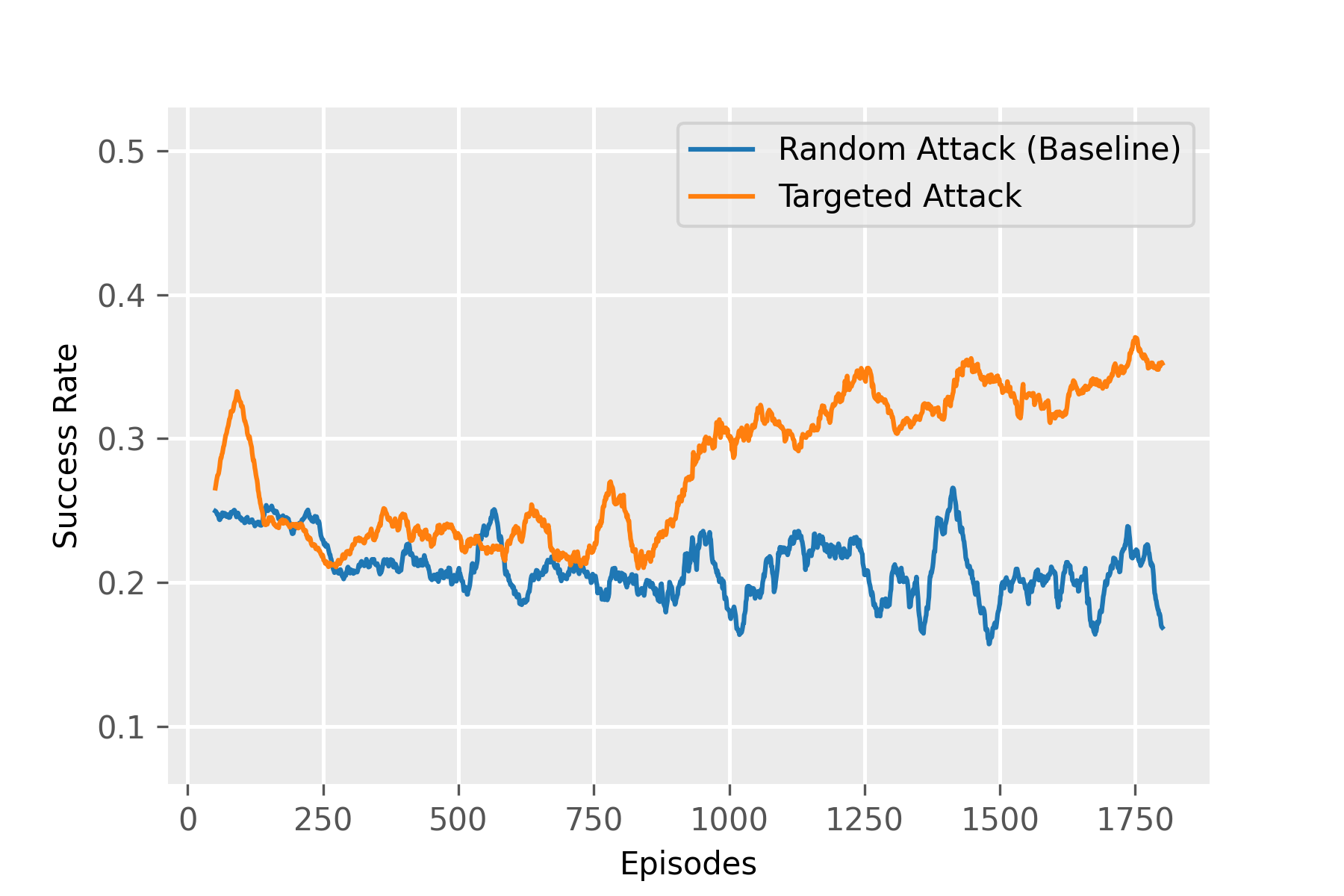}
        \caption{Success Rate Comparison on Atari Breakout.}
    \end{subfigure}
    \caption{Success Rate of Reward Shifting Attack. }
    \label{fig:drop_shift_SR}
\end{figure*}

Finally, we investigate the potency of the reward shifting attack in the context of the targeted attack. We report those results in Figure~\ref{fig:drop_shift_SR}. We observe that the targeted reward shifting attack actually yields a higher success rate compared to the baseline random reward shifting attack in both Atari Pong and Breakout environments, with the success rate gap increasing with the number of episodes.  
\vspace{-12pt}

\section{Conclusions}
\vspace{-2pt}
We study the problem of reward delay attacks against reinforcement learning agents. Our empirical findings suggest that it is possible to induce a sub-optimal policy, or even a specific target policy by strategically reshuffling the true reward sequence. 
Indeed, we find that even randomly shuffling rewards within relatively short time intervals is already sufficient to cause learning failure.
This raises a potentially serious security threat to different downstream applications that rely on RL. 
Moreover, we showed that reward shifting attacks that assure that reward signals are not observed out of order, also have a disastrous effect on DQN learning. Our finding shows that current deep RL training strategy falls far short of ensuring adequate robustness to these in many cases.

A natural subject for future work is to develop mitigation techniques that can assure adequate synchrony in reinforcement learning when it is necessary.
Often, mitigations of this kind must involve hardware support that enables us to assure synchrony of state and reward information.
A natural second open question is whether it is possible to avoid strong reliance on such hardware support by developing reinforcement learning approaches that have weaker synchrony requirements.

\paragraph{Acknowledgments} This research was supported in part by the National Science Foundation (grants IIS-1905558, IIS-2214141, ECCS-2020289),  Army Research Office (grant W911NF1910241), and NVIDIA.

\bibliographystyle{splncs04}
\bibliography{main}

\newpage

\section{Appendix}
In this section, we present the end-to-end algorithmic approach of reward delay attack strategy and reward shifting attack strategy in algorithm \ref{alg:2} and \ref{alg:3} respectively.
\vspace{-10pt}
\begin{algorithm}[!ht]  
  \caption{ \textbf{Reward Delay Attack Strategy} }  
  \label{alg:2}  
  \begin{algorithmic}[1]  
    \Require 
        $Q_{t}(s_t,a_t,\theta)$ - learner's ($\mathcal{A}$) current Q-network parameterized by $\theta$;
        An attacker agent $\mathcal{A}^{\prime}$ with corresponding MDP $\mathcal{M}^{\prime}=(\Xi^{\prime},\rho^{\prime},\tau^{\prime})$, Q-network($Q^{\prime}$) parameterized by $\theta^{\prime}$, disk $\mathcal{D}_{0}$ to store the rewards and attack constraint parameters $\{\delta, d \}$; Action set of the $\mathcal{A}^{\prime}$, i.e. A$\in$\{1,2,..,d\} where $d$ is the disk size; T - No. of time-steps in an episode; $\mathcal{B}$, $\mathcal{B}^{\prime}$ - Replay buffer of $\mathcal{A}$ and $\mathcal{A}^{\prime}$; 
    \For {t=0, 1, ..., T}
        \State $\mathcal{A}$ interacts with the environment by taking an action $a_t$ given a state $s_t$. Environment transits according to $P(s_{t+1}|s_t,a_t)$, and feeds back a reward signal $r_t$ and a next state $s_{t+1}$.
        \State $\mathcal{A}^{\prime}$ extracts ($s_t,a_t,r_t,s_{t+1}$) and push $r_t$ into $\mathcal{D}_{t}$.
        \State $\mathcal{A}^{\prime}$ leverages the following information to act:
        $\emph{info}_{t}$ = $[s_t, a_t, r_t, \mathcal{D}_t]$ and learner's  $Q_t$.
        \State $\mathcal{A}^{\prime}$ takes action $a^{\prime}_t$ according to $\epsilon$-greedy behavior policy 
        $$
        \emph{i}\xleftarrow{}\left\{
        \begin{aligned}
            argmax_{a\in A}\>\>Q^{\prime}_t(\emph{info}_{t},a,\theta^{\prime}),\ \ \ \ \ \ \ \text{If}\ 1-\epsilon\\
            \text{Randomly select from } A, \ \ \ \ \ \ \ \text{else}
        \end{aligned}
        \right\}
        $$
        \State $r_t$ (the true reward) is exchanged with $r^{\mathcal{D}}_{i}$(The reward stored at the i'th index in the disk $\mathcal{D}_{t}$). Accordingly, t'th transitional tuple data is updated as ($s_t,a_t,r^{\mathcal{D}}_i,s_{t+1}$) and stored into $\mathcal{B}$.
        \State Disk $\mathcal{D}_t$ is updated by removing $r^{\mathcal{D}}_i$ from the disk. We also \emph{drop} the reward from $\mathcal{D}_t$ whose stored-time (i.e. delay duration) exceeds $\delta$.
        \State $\mathcal{A}$ performs Q-learning given the poisoned mini-batch transition data sampled from $\mathcal{B}$, where t'th transition tuple of the mini-batch is represented as $(s_{t},a_{t},r^{\mathcal{D}}_{i},s_{t+1})$:
            $$
            y_j\xleftarrow{}\left\{
            \begin{aligned}
                r^{\mathcal{D}}_{i},\ \ \ \  \ \text{If}\ s_{t+1} \ \text{is a Terminal State}\\
                r^{\mathcal{D}}_{i}+max_{a^\prime}(Q_{t}(s_{t+1},a^{\prime},\theta)), \     \ \ \ \text{else}
            \end{aligned}
            \right.
            $$
            $\>\>\>\>\>\>$ Update the learner's ($\mathcal{A})$ Q-network parameters from $Q_{t}$ to $Q_{t+1}$ by minimizing the loss $\mathcal{L}(\theta) = (y_{j} - Q_{t}(s_{t},a_{t},\theta) )^{2}$ following DQN update rule. 
        \State $\mathcal{A}^{\prime}$ computes reward ${r}^{\textbf{attacker}}_{t}$ following the methods as described in section \ref{sec:algo} and stores the transition tuple $(\emph{info}_{t},a^{\prime}_{t},r^{\textbf{attacker}}_{t}, \emph{info}_{t+1})$ into its own replay buffer $\mathcal{B}^{\prime}$.
        \State $\mathcal{A}^{\prime}$ performs Q-learning given the attacker's mini-batch transition data sampled from $\mathcal{B}^{\prime}$, where t'th transition tuple is denoted as  $(\emph{info}_{t},a^{\prime}_{t},r^{\textbf{attacker}}_{t}, \emph{info}_{t+1})$:
            $$
            y_j\xleftarrow{}\left\{
            \begin{aligned}
                {r}^{\textbf{attacker}}_{t},\ \ \ \  \ \text{If}\ \emph{info}_{t+1} \ \text{is a Terminal State}\\
                {r}^{\textbf{attacker}}_{t}+max_{a^\prime \in A}(Q^{\prime}_{t}(\emph{info}_{t+1},a^{\prime},\theta^{\prime})), \     \ \ \ \text{else}
            \end{aligned}
            \right.
            $$
        \State $\mathcal {A}^{\prime}$ also updates it's Q-network parameter $\theta^{\prime}$ by minimizing $\mathcal{L}(\theta) = (y_{j} - Q^{\prime}_{t}(\emph{info}_{t},a^{\prime}_{t},\theta^{\prime}) )^{2}$.
        
   \EndFor
    \State \Return $\mathcal{A}$ and it's corresponding Q-network parameter $\theta$;  
  \end{algorithmic}  
\end{algorithm} 

\begin{algorithm}[!ht]  
  \caption{ \textbf{Reward Shifting Attack Strategy} }  
  \label{alg:3}  
  \begin{algorithmic}[1]  
    \Require 
        $Q_{t}(s_t,a_t,\theta)$ - learner's ($\mathcal{A}$) current Q-network parameterized by $\theta$;
        An attacker ($\mathcal{A}^{\prime}$) with corresponding MDP $\mathcal{M}^{\prime}=(\Xi^{\prime},\rho^{\prime},\tau^{\prime})$, Q-network($Q^{\prime}$) parameterized by $\theta^{\prime}$, disk $\mathcal{D}_{0}$ to store the rewards and attack constraint parameters $\{l, d \}$; Action set of the $\mathcal{A}^{\prime}$, i.e. A$\in$\{1,2,..,K\}; $\mathcal{B}, \mathcal{B}^{\prime}$ - are Learner's and attacker's Replay Buffer respectively; T - The no. of time-steps in an episode.
    \For {t=0, 1, ..., T}
        \State $\mathcal{A}$ interacts with the environment which results a transition tuple ($s_t,a_t,r_t,s_{t+1}$).
        \State $\mathcal{A}^{\prime}$ Extract ($s_t,a_t,r_t,s_{t+1}$) and push reward $r_t$ into $\mathcal{D}_{t}$.
        \State $\mathcal{A}^{\prime}$ leverages the following information to act when $\mathcal{D}_{t}$ is full: 
        $\emph{info}_{t}$ = $[s_t, a_t, r_t, \mathcal{D}_t]$ and $Q_t$.
        \State $\mathcal{A}^{\prime}$ takes action $a^{\prime}_t$ according to $\epsilon$-greedy behavior policy when $\mathcal{D}_t$ is full.
        $$
        \emph{i}\xleftarrow{}\left\{
        \begin{aligned}
            argmax_{a\in A}\>\>Q^{\prime}_t(\emph{info}_{t},a,\theta^{\prime}),\ \ \ \ \ \ \ \text{If}\ 1-\epsilon\\
            \text{Randomly select from } A, \ \ \ \ \ \ \ \text{else}
        \end{aligned}
        \right\}
        $$
        \State $\mathcal{A}^{\prime}$ publishes an updated reward sequence with the following modifications: Firstly, the rewards stored from index $0$ to $i$ are dropped, Secondly, the reward sequence ranging from index $(i+1)$ to $d$ is published in sequence. We denote the attacker's published reward sequence as $\{r^{\mathcal{D}}_{1}, r^{\mathcal{D}}_{2},..,r^{\mathcal{D}}_{t},..\}$.  
        \State $\mathcal{A}$ performs Q-learning given the poisoned mini-batch transition data sampled from $\mathcal{B}$, where t'th transition tuple of the mini-batch is represented as $(s_{t},a_{t},r^{\mathcal{D}}_{t},s_{t+1})$:
            $$
            y_j\xleftarrow{}\left\{
            \begin{aligned}
                r^{\mathcal{D}}_{t},\ \ \ \  \ \text{If}\ s_{t+1} \ \text{is a Terminal State}\\
                r^{\mathcal{D}}_{t}+max_{a^\prime}(Q_{t}(s_{t+1},a^{\prime},\theta)), \     \ \ \ \text{else}
            \end{aligned}
            \right.
            $$
            $\>\>\>\>\>\>$ Update the learner's ($\mathcal{A})$ Q-network parameters from $Q_{t}$ to $Q_{t+1}$ following DQN update rule.  
        \State $\mathcal{A}^{\prime}$ computes reward ${r}^{\textbf{attacker}}_{t}$ following the methods as described in section \ref{sec:algo} and stores the transition tuple $(\emph{info}_{t},a^{\prime}_{t},r^{\textbf{attacker}}_{t}, \emph{info}_{t+1})$ into its own replay buffer $\mathcal{B}^{\prime}$.
        \State $\mathcal{A}^{\prime}$ performs Q-learning given the attacker's mini-batch transition data sampled from $\mathcal{B}^{\prime}$, where t'th transition tuple is denoted as  $(\emph{info}_{t},a^{\prime}_{t},r^{\textbf{attacker}}_{t}, \emph{info}_{t+1})$:
            $$
            y_j\xleftarrow{}\left\{
            \begin{aligned}
                {r}^{\textbf{attacker}}_{t},\ \ \ \  \ \text{If}\ \emph{info}_{t+1} \ \text{is a Terminal State}\\
                {r}^{\textbf{attacker}}_{t}+max_{a^\prime \in A}(Q^{\prime}_{t}(\emph{info}_{t+1},a^{\prime},\theta^{\prime})), \     \ \ \ \text{else}
            \end{aligned}
            \right.
            $$
        \State $\mathcal {A}^{\prime}$ also updates it's Q-network parameter $\theta^{\prime}$ by minimizing $\mathcal{L}(\theta) = (y_{j} - Q^{\prime}_{t}(\emph{info}_{t},a^{\prime}_{t},\theta^{\prime}) )^{2}$.
        
   \EndFor
    \State \Return $\mathcal{A}$ and it's corresponding Q-network parameter $\theta$;  
  \end{algorithmic}  
\end{algorithm} 

\end{document}